\documentclass{article} %
\usepackage{iclr2027_conference,times}

\usepackage{amsmath,amsfonts,bm}

\def\eqref#1{equation~\ref{#1}}

\def\1{\bm{1}}

\def\vtheta{{\bm{\theta}}}
\def\vphi{{\bm{\phi}}}

\def\vu{{\bm{u}}}
\def\vv{{\bm{v}}}

\def\vx{{\bm{x}}}

\def\mA{{\bm{A}}}

\def\mI{{\bm{I}}}

\def\mM{{\bm{M}}}

\def\mP{{\bm{P}}}

\def\mU{{\bm{U}}}

\DeclareMathAlphabet{\mathsfit}{\encodingdefault}{\sfdefault}{m}{sl}
\SetMathAlphabet{\mathsfit}{bold}{\encodingdefault}{\sfdefault}{bx}{n}

\newcommand{\E}{\mathbb{E}}

\newcommand{\R}{\mathbb{R}}

\let\originalleft\left
\let\originalright\right
\renewcommand{\left}{\mathopen{}\mathclose\bgroup\originalleft}
\renewcommand{\right}{\aftergroup\egroup\originalright}

\newcommand{\diff}{\mathop{}\!\mathrm{d}}
\let\originalpartial\partial
\renewcommand{\partial}{\mathop{}\!\mathrm{\originalpartial}}

\newcommand{\T}{^\mathrm{T}}

\usepackage{hyperref}
\usepackage{url}
\usepackage{graphicx}
\usepackage{caption}
\usepackage{booktabs}       %
\usepackage[table]{xcolor} %
\usepackage{multirow}
\usepackage{mathtools}
\usepackage{xspace}
\usepackage{listings}
\usepackage{enumitem}
\usepackage{wrapfig}

\title{Rethinking the design space of Vanilla Pixel Diffusion Transformers}

\author{
Soumik Mukhopadhyay$^{1,2}$%
\thanks{Work done while an intern at Adobe Research.}
\quad
Long Mai$^{2}$
\quad
Abhinav Shrivastava$^{1}$
\quad
Aniruddha Mahapatra$^{2}$%
\thanks{Project lead.}
\\[2pt]
$^1$~University of Maryland, College Park
\quad
$^2$~Adobe Research
}

\makeatletter
\DeclareRobustCommand\onedot{\futurelet\@let@token\@onedot}
\def\@onedot{\ifx\@let@token.\else.\null\fi\xspace}

\def\eg{\emph{e.g}\onedot} 
\def\ie{\emph{i.e}\onedot}

\newcommand{\vpred}{\ensuremath{\vu}\xspace}
\newcommand{\xpred}{\ensuremath{\vx_0}\xspace}
\newcommand{\method}{\textsc{MFlow}\xspace}

\iclrfinalcopy %
\begin{document}

\maketitle

\begin{figure}[h]
\vspace{-1em}
\begin{center}
\includegraphics[width=\linewidth]{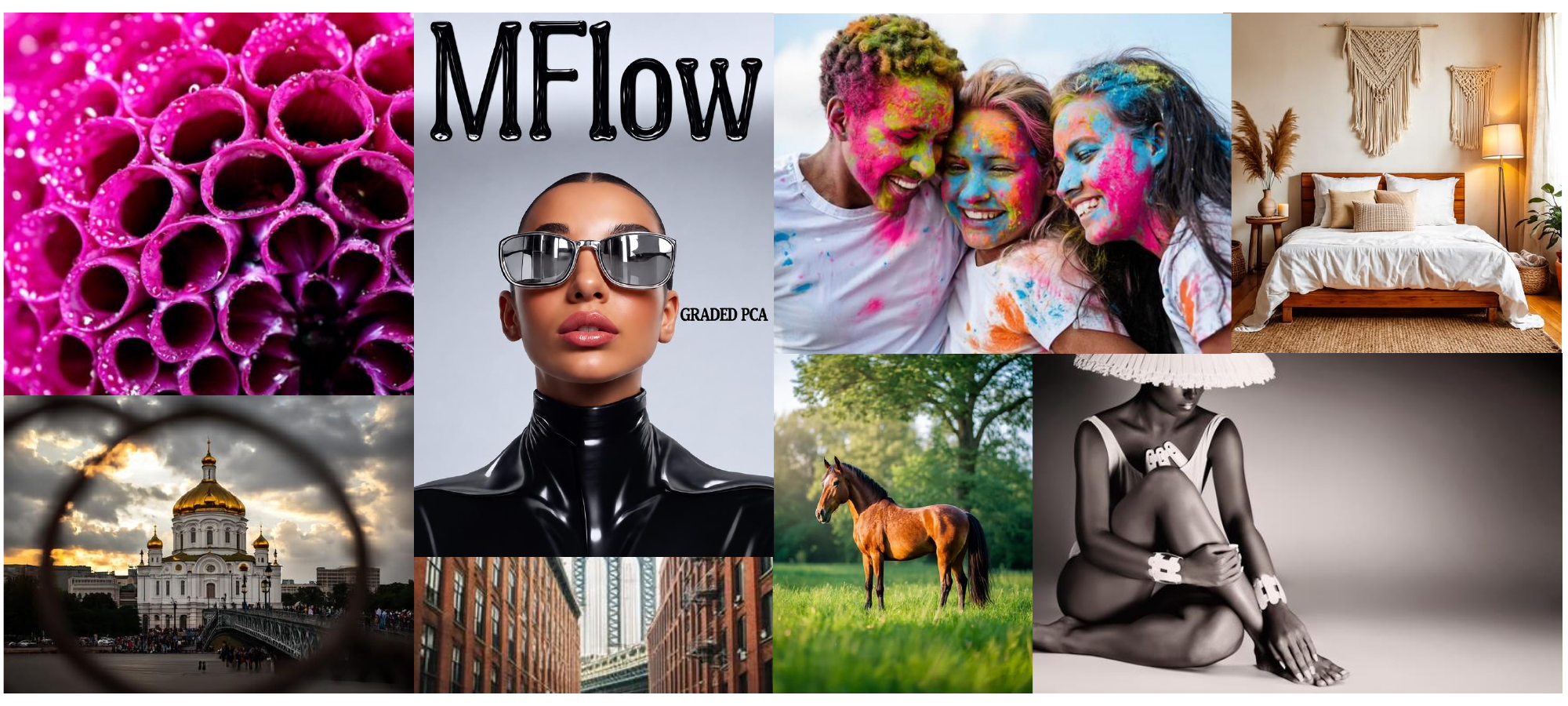}
\end{center}
\vspace{-1em}
\caption{Sample 512px text-to-image generations with our \method-XXL/16 model. 
}
\end{figure}

\begin{abstract}

Pixel-space diffusion models have recently re-emerged as a compelling alternative to latent diffusion, avoiding the lossy latent bottleneck and separately trained autoencoder. JiT shows that a vanilla diffusion transformer can generate directly in pixel space by replacing the commonly used velocity prediction (\vpred) with clean image prediction (\xpred). In this work, we revisit this choice and introduce \method, an adaptive parameterization that learns to interpolate between \xpred and \vpred prediction across principal directions of the data, which can be learned end-to-end with diffusion model training. We find that the preferred parameterization varies systematically with the data representation and model capacity. Notably, increasing model capacity shifts preference toward \vpred prediction from \xpred prediction. We evaluate \method across ImageNet, UCF-101 video, and large-scale text-to-image generation, demonstrating competitive or improved generative performance across diverse settings.
In addition, we find that vanilla pixel diffusion transformers, irrespective of parameterization, retain a persistent high-frequency deficit relative to real images. To address this, we introduce \emph{ConvRefiner}, a lightweight convolutional module trained post-hoc with adversarial loss while keeping the diffusion backbone frozen. With only a small amount of additional training, \emph{ConvRefiner} restores the high-frequency statistics of generated images while preserving FID.
Project webpage: \href{https://soumik-kanad.github.io/mflow-diffusion}{soumik-kanad.github.io/mflow-diffusion}.
\end{abstract}

\section{Introduction}

Diffusion and flow-based models have become a leading paradigm for visual generation in recent years.
Modern versions of these models predominantly operate in the latent space, obtained from compressing the pixel-space data with a separately-trained autoencoder. 
However, latent diffusion models (LDMs) (\cite{ldm, dit, sit, sd3, yu2025repa, singh2026improved}) introduce their own limitations -- a lossy bottleneck leads to unfaithful reconstruction after compression, and there is a need for a separately trained compression autoencoder,  which is typically unaware of the downstream generative objective.

Recent works (\cite{jit,dip,pixeldit,ssd,chen2025pixelflow,DeCo,
pixelgen,minit2i2026}) have revived diffusion directly in pixel space for image generation, removing the
aforementioned drawbacks in LDMs.
In particular, JiT (\cite{jit}), demonstrates that vanilla diffusion transformers can achieve competitive generative performance directly on raw pixels, eliminating the need for pretrained autoencoders or complicated architectural modifications.
Its aggressive patchification keeps the
sequence length manageable, \eg, its \(16\times16\) patchification produces
equivalent number of tokens as a \(16\times\) spatial compression VAE followed
by \(1\times1\) latent patchification. 
At the same time, such aggressive patchification poses a challenge in pixel-space diffusion as each token must
model a high-dimensional block of raw pixels, raising two questions:
\emph{what prediction parameterization is best suited to such high-dimensional
patches}, and \emph{can a vanilla large-patch transformer faithfully model
fine-scale image details?} In this work, we aim to improve the 
pixel-space diffusion paradigm along these axes.

JiT addresses the first question with clean-image (\xpred) prediction.
Under the manifold assumption, clean data lies near a low-dimensional
manifold, whereas noise and velocity span the ambient space, making
\xpred easier for an undercomplete network to predict. However, sampling requires converting \xpred to \vpred (velocity), which becomes ill-conditioned near the low-noise regime. 
AsymFlow (\cite{asym}) reduces this problem by predicting
\vpred in a selected rank-\(r\) PCA subspace and \xpred in its
complement, keeping the prediction target effectively in a low dimensionality. However, this hard binary cutoff rank must be chosen for each setting through  expensive training sweeps, \eg, AsymFlow needed 6 trainings to tune the rank-\(r\).
Where as \cite{kdiff} use a learnable but globally shared interpolation between \xpred and \vpred, not reducing the effective noise dimensionality.

We introduce \textbf{\method}, an adaptive parameterization with a
continuous gain \(m_i\in[0,1]\) for every principal direction of data. Notably, this differentiable velocity-recovery formulation
lets us jointly train the generator and gains with the standard
velocity-space flow-matching loss, letting model learn the best combination of \vpred and \xpred per principal component. \method therefore
replaces the AsymFlow's manually selected rank and binary mask with a
continuous parameterization that adapts to the data representation,
objective, and model capacity.
Our experiments show that \method can be adapted not only to pixel-space image generation, but also to diverse paradigms, like pixel-space video generation, text-to-image generation at multiple model capacities, and high-dimensional latents like RAEv2 (\cite{singh2026improved}). Further visualization of the learned
direction-wise gains as \(\mM\)-curves, a plot of \(m_i\)'s in decreasing direction variance order, reveal interesting insights, such as increasing model capacity
shifts more directions toward velocity-like prediction.

Prediction parameterization alone does not resolve all limitations of
vanilla pixel diffusion transformers. Irrespective of the parameterization,
we find that JiT-family models exhibit a persistent high-frequency deficit
relative to real images. This suggests a limitation
of the minimalist large-patch design, where patch tokens are projected
directly back to pixels using \texttt{MLP}-layers without an explicit local refinement pathway. We
 introduce \textbf{\emph{ConvRefiner}}, a lightweight residual convolutional
module trained post hoc with perceptual and adversarial supervision while
keeping the diffusion backbone frozen. With only 5\% post-hoc training,
\emph{ConvRefiner} brings the generated high-frequency spectrum closer to that of
real images while preserving FID.
Overall, this work presents adaptive prediction parameterization and fine-scale signal modeling 
as complementary directions for improving vanilla pixel diffusion transformers.

To summarize, the key contributions are:
\begin{itemize}[noitemsep,topsep=2pt,leftmargin=2em]
    \item We introduce \method, a framework of continuous, direction-adaptive prediction
    parameterization whose gains are learned jointly with the generator
    under a standard velocity-space flow-matching objective. We show \method is a general framework and can be used for diverse domains like pixel-space video generation, high-dimensional latents (like RAEv2) in addition to pixel-space image generation.
    
    \item We identify a high-frequency deficit in vanilla
    JiT-style models and introduce \emph{ConvRefiner}, a lightweight post-hoc
    module that restores high-frequency details while preserving fidelity matching the frequency spectrum of real images.
\end{itemize}

\newpage
\begin{figure}[h]
\begin{center}
\includegraphics[width=\linewidth]{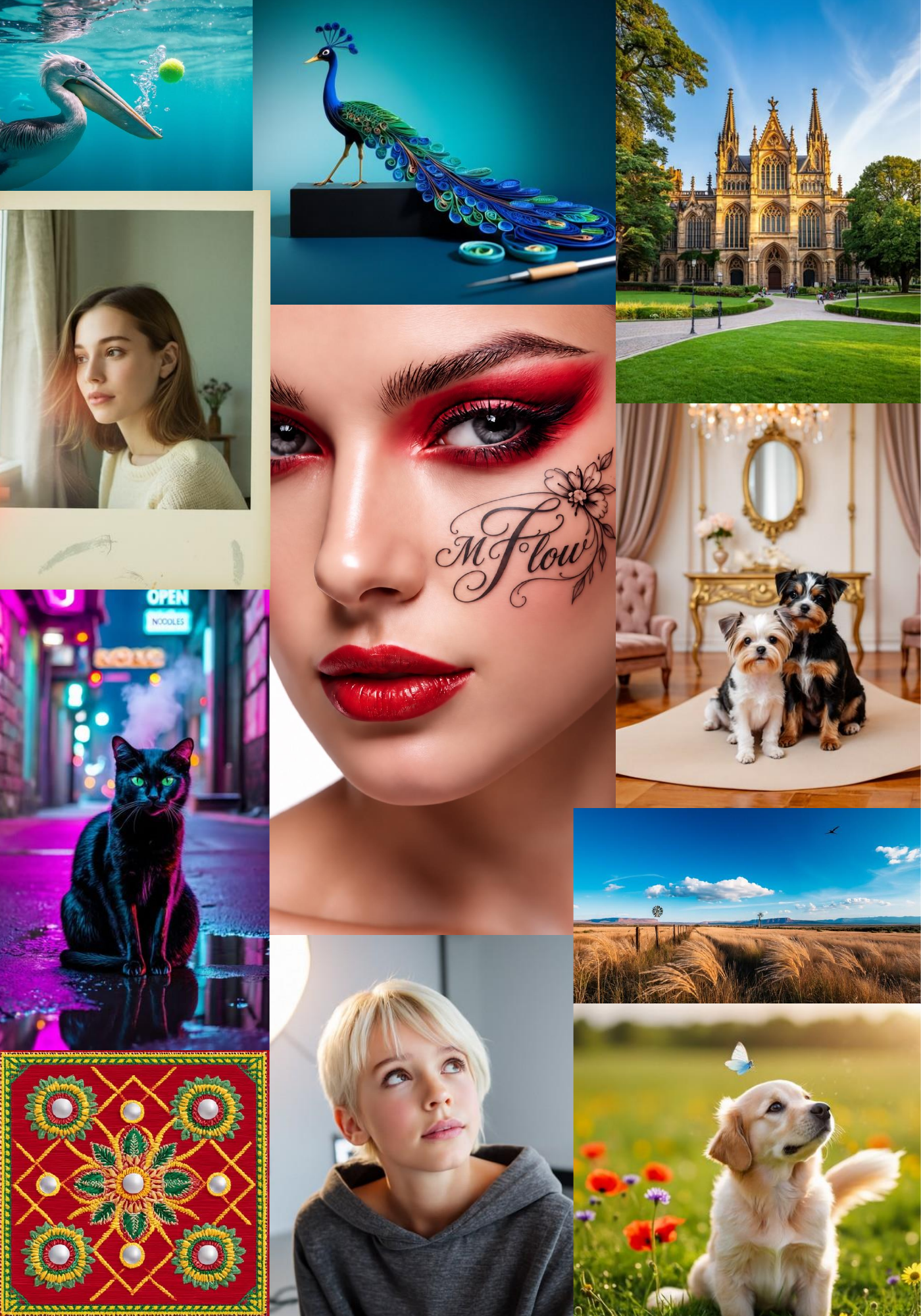}
\end{center}
\caption{\textbf{Additional Text-to-Image Results.} Sample 512px text-to-image generations with our MFLOW-XXL/16 model.}
\label{fig:imagenet_qual}
\end{figure}
\newpage

\section{Related Work}

\noindent\textbf{Pixel-space diffusion and prediction parameterization.}
Early diffusion models generated directly in pixel space (\cite{ddpm,iddpm,adm,edm}), while subsequent works improved the scalability of end-to-end pixel generation through architectural and formulation advances (\cite{rin,hoogeboom2023simple,hoogeboom2025simpler,hdit,chen2025pixelflow,ssd}). Recent approaches have renewed interest in pixel-space generation using vanilla or specialized transformer architectures (\cite{jit,dip,pixeldit,guo2026pixelu}), with recent studies further exploring its scaling to text-to-image generation (\cite{jiang2026empiricalstudytrainingpixelspace,minit2i2026, DeCo, pixeldit, pixelgen, chen2025pixelflow}). In the context of our parameterization study, JiT (\cite{jit}) advocates clean-image prediction for high-dimensional pixel modeling, an approach also adopted by recent large-scale pixel-space models such as Tuna-2 (\cite{liu2026tuna2pixelembeddingsbeat}). k-Diff~\cite{kdiff} and AsymFlow~\cite{asym} further explore adaptive combinations of clean-data and velocity prediction. Instead of deciding a predefined fixed formulation, which can vary across architectures and datasets, \method lets the model learn the best fit between clean image and velocity prediction.

\noindent\textbf{Details and frequency modeling.}
Several recent works study the difficulty of modeling fine-scale information in
vanilla pixel-space diffusion transformers with large spatial patch sizes, such as
\(16\times16\)--\(32\times32\). Existing approaches use perceptual
supervision (\cite{pixelgen}), frequency-aware objectives or
flows (\cite{fvloss,wait,spectralforcing}), frequency-specialized
architectures (\cite{DeCo,ma2026frequencyboosterfullfrequencymodelinghighfidelity}),
or improved information pathways (\cite{guo2026pixelu}). Concurrent work also shows
that a frozen JiT model can recover additional high-frequency detail through
self-guidance (\cite{fu2026frozen}). Although some of these approaches also improve
training efficiency (\cite{fvloss,DeCo}), they generally introduce specialized
architectural components or modify the training objective, and their generated
frequency statistics can still exhibit a high-frequency gap relative to real
images (\cite{fvloss,DeCo,wait}). In contrast, our lightweight \emph{ConvRefiner} is trained
post hoc while keeping the pretrained diffusion backbone frozen. It recovers the
missing high-frequency statistics without modifying the backbone or its original
training procedure, while preserving FID.

\section{Model Prediction Parameterization}
\subsection{Motivation}
\label{sec:param_motivation}
\noindent\textbf{Flow Matching.} In the linear stochastic interpolant/rectified-flow formulation used by SiT (\cite{sit}) and JiT (\cite{jit}), for signal $\vx_0 \sim p_{\text{data}}$ and noise $\bm{\epsilon} \sim \mathcal{N}(\bm{0}, \mI)$, both in $\R^D$, the forward interpolation at timestep $t$: 
$\vx_t \;=\; t\,\vx_0 + (1-t)\,\bm{\epsilon}$.
The generative process is the probability-flow ODE that transports pure noise $\bm{\epsilon}$ ($t=0$) to clean signal $\vx_0$ ($t=1$) along the conditional expectation of the interpolant's time derivative, the velocity field $\vu(\vx_t,t)=\E_{}\left[\frac{\diff\vx_t}{\diff t} \middle| \vx_t\right]=\E_{}\left[\vx_0-\bm{\epsilon} \middle| \vx_t\right]$.
A network $f_\vtheta(\vx_t,t)$ is trained by regressing a target derived from
$(\vx_0,\bm{\epsilon},t)$.

\noindent\textbf{Adaptive parameterization.} Common non-adaptive parameterization/regression-targets include: noise $\bm{\epsilon}$-prediction, velocity $\vu$-prediction , and clean signal $\vx_0$-prediction. These are algebraically equivalent but differ in conditioning. EDM (\cite{edm}) introduces an input/output preconditioning, where the network effectively predicts a noise-level dependent  combination of clean data and noise. However, JiT (\cite{jit}) demonstrates that such a preconditioning substantially underperforms $\vx_0$-prediction for high-dimensional pixel patches using a vanilla transformer backbone, motivating the use of standard clean image prediction. More recently, k-Diff (\cite{kdiff}) introduces a learnable scalar interpolation between prediction targets, showing that the preferred parameterization varies across pixel and latent representations. However, its globally shared coefficient cannot account for directional differences within a representation. 

AsymFlow (\cite{asym}) addresses this limitation by assigning velocity $\vu$-prediction to a selected low-rank subspace while clean image $\vx_0$-prediction in its complement.
More concretely, let
$\mU=[\vv_1,\dots,\vv_D]$ be the orthonormal patch-PCA eigenvectors of the data (variances
$\lambda_1\ge\dots\ge\lambda_D$), and let $\mA=\mU_{:,1:r}$ be the top-$r$ directions  with $\mA\T\mA = \mI_r$, then the orthogonal projector $\mP \;=\; \mA\mA\T$ has the properties $\mP^2=\mP $ and $(\mI-\mP)\mP=\bm 0$. Their model predicts the asymmetric target $ \vu_\mathrm{A} \;=\; \vx_0 - \mP\bm{\epsilon}$, 
\ie, 
predict the full clean signal, but only the low-rank projection of the noise.
Because $\mP$ is
a projector, the target splits cleanly into the velocity inside the subspace $\mP\,\vu_\mathrm{A} = \mP(\vx_0-\bm{\epsilon})=\mP\,\vu$, and the clean data on the complement $(\mI-\mP)\,\vu_\mathrm{A} = (\mI-\mP)\vx_0$. So each PCA direction is treated either as velocity ($i\le r$) or as $\vx_0$ ($i>r$). The full velocity is recovered from $\vu_\mathrm{A}$ and $\vx_t$ by: $\vu \;=\; \mP\,\vu_\mathrm{A} \;+\; (\mI-\mP)\,\frac{\vx_t + \vu_\mathrm{A}}{\,1-t\,}$.

\noindent\textbf{Limitations of previous adaptive parameterizations.} \cite{asym} note that the default $\vx_0$-prediction and k-Diff parameterization are less expressive.
On the other hand, AsymFlow makes hard-cutoff decision at a fixed rank $r$, and this chosen subspace is dataset/modality dependent, requiring an expensive hyperparameter search over multiple parameterization setting.
Furthermore, restricting the noise term to the top-$r$ (highest-variance) directions is arbitrary -- \eg, we found that $\mathrm{PCA}[9\!:\!16]$ can beat $\mathrm{PCA}[1\!:\!8]$ (FID $3.54$ vs $3.59$ on the B-level model trained on ImageNet). 
Finally, the binary nature of the cutoff, while useful for interpretation, is not necessary: it extends naturally to a continuous, per-direction weighting that softly mixes the clean-signal ($\vx_0$) and velocity ($\vv$) prediction. Building on these insights, we introduce \method, which learns a continuous, direction-dependent parameterization, allowing the prediction target to adapt across directions without requiring a hard subspace partition or costly search for parameterization.

\begin{figure}[t]
\begin{center}
\includegraphics[width=\linewidth]{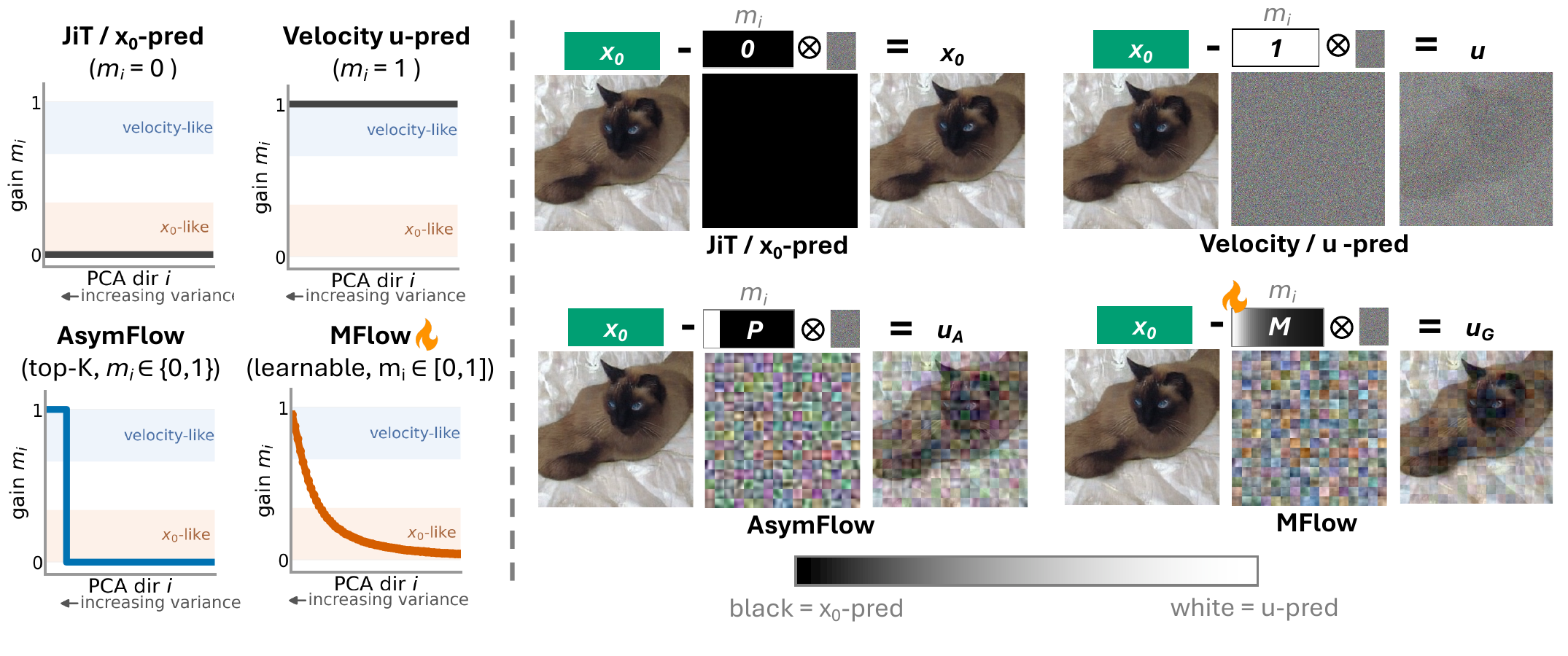}
\end{center}
\caption{\textbf{Overview of \method Formulation.} We propose a principal data direction-wise learnable parameterization as a generalized framework in contrast with previous fixed parameterizations.
}
\label{fig:mflow_overview}
\end{figure}
\subsection{\method: Direction-Adaptive Learnable Reparameterization}
\label{sec:mflow_formulation}
Instead of a hard cutoff, we assign each spectral direction a soft, learnable weight in $[0,1]$. Replace the binary projector $\mP$ by a symmetric graded operator:
\begin{equation}
    \mM \;=\; \mU\,\mathrm{diag}(m_1,\dots,m_D)\,\mU\T,\qquad m_i\in[0,1],\qquad \mM \vv_i = m_i \vv_i, \label{eq:mgrad}
\end{equation}
with the graded target (as shown in Fig.~\ref{fig:mflow_overview}):
\begin{equation}
    \,\vu_\mathrm{G} \;=\; \vx_0 - \mM\bm{\epsilon}\,
    . \label{eq:mflow_target}
\end{equation}
$\mM$ is generally not a projector ($\mM^2\ne\mM$ unless every $m_i\in\{0,1\}$), so it does
not split $\R^D$ into the low-rank subspace $\mathrm{Im}(\mP)$ and its orthogonal complement $\mathrm{Im}(\mI-\mP)$. Instead every direction carries its own
continuous velocity-data mixing weight.

\noindent\textbf{Component-wise (PCA basis).}
Let $\tilde{\vx}\coloneqq\mU\T\vx$ denote the transformation of $\vx$ in the PCA basis, then we have the $i^{\text{th}}$ component $\tilde x_i=\vv_i\T\vx$. Because $\mM$ is diagonal in this basis, $(\widetilde{\mM\vx})_i = m_i\tilde x_i$, and the graded target decouples per direction:
\vspace{-0.5em}
\begin{equation}
    \tilde u_{\mathrm{G},i} \;=\; \tilde x_{0,i} - m_i\,\tilde\epsilon_i \label{eq:mflow_comp}
\end{equation}
where the directional velocity is $\tilde u_i = \tilde x_{0,i}-\tilde\epsilon_i$. Here $m_i$ denotes how velocity-like the target is in this direction: $m_i=1\Rightarrow \vu\text{-pred}$ while $m_i=0\Rightarrow \vx_0\text{-pred}$. \method therefore strictly generalizes both the endpoints as well as AsymFlow ($m_i=\mathbf 1[i\le r]$). 

\noindent\textbf{Choosing the gains.}
We initialize the gains with $m_i=\frac{\hat\lambda_i}{\hat\lambda_i+c}$ where 
$\hat\lambda_i=\lambda_i/\lambda_{\max}$ is the normalized variance, and $c$ is a small constant. By default these gains are learnable $m_i=\mathrm{sigmoid}(\phi_i)$ with a learnable
per-direction logit $\phi_i$.

\noindent\textbf{Exact velocity recovery.}
The forward  
$\vx_t \;=\; t\,\vx_0 + (1-t)\,\bm{\epsilon}$
and
target Eq.~(\ref{eq:mflow_comp}) give two linear equations in the two unknowns$(\tilde x_{0,i},\tilde\epsilon_i)$ in each eigendirection:
\begin{equation}
    \begin{bmatrix}\tilde x_{t,i}\\[2pt] \tilde u_{\mathrm{G},i}\end{bmatrix}
    =
    \begin{bmatrix} t & 1-t\\[2pt] 1 & -m_i \end{bmatrix}
    \begin{bmatrix}\tilde x_{0,i}\\[2pt] \tilde\epsilon_i\end{bmatrix},
    \label{eq:mflow_sys}
\end{equation}
    The system matrix has determinant $-\Delta_i$ with $\Delta_i \;=\; (1-t)+m_i t \ge 1-t > 0$; since it is nonzero (strictly negative) for every $m_i\in[0,1]$ 
    , the solve is always exact:
\begin{equation}
    \;\tilde x_{0,i} \;=\; \frac{(1-t)\,\tilde u_{\mathrm{G},i} + m_i\,\tilde x_{t,i}}{\Delta_i}\;
    ,
    \qquad
    \tilde\epsilon_i \;=\; \frac{\tilde x_{t,i} - t\,\tilde u_{\mathrm{G},i}}{\Delta_i},
    \label{eq:mflow_recover_comp}
\end{equation}
Equivalently, in the operator form the $vx_0$ and $vu$ can be recovered from the target $\vu_\mathrm{G}$ using:
\begin{equation}
    \begin{aligned}
    \vx_0 &\;=\; \big[t\,\mM + (1-t)\,\mI\big]^{-1}\big[(1-t)\,\vu_\mathrm{G} + \mM\,\vx_t\big],\\
    \vu &\;=\; \frac{\vx_0 - \vx_t}{1-t} \;=\; \big[t\,\mM + (1-t)\,\mI\big]^{-1}\big[\vu_\mathrm{G}-(\mI-\mM)\,\vx_t\big].
    \end{aligned}
    \label{eq:mflow_recover_op}
\end{equation}
Setting $m_i\in\{0,1\}$ reduces Eq~(\ref{eq:mflow_recover_op}) exactly
to the AsymFlow recovery rule.

\noindent\textbf{Training and sampling.} $\mM$ changes neither the forward process nor the
transport target $\vu$, only the coordinate-wise conditioning of the
network's regression target. The network outputs $\hat{\vu}_G$ as the learned approximation for target $\vu_\mathrm{G}$ which we map it to the
full velocity $\hat{\vu}$ via Eq~(\ref{eq:mflow_recover_op}) and train a velocity-space flow-matching loss,
\begin{equation}
    \mathcal L(\vtheta)=\E_{\vx_0,\bm{\epsilon},t}\big[\,w(t)\,\|\hat{\vu}(\hat{\vu}_G;\vx_t,t)-(\vx_0-\bm{\epsilon})\|_2^2\,\big]. \label{eq:mflow_loss}
\end{equation}
Additionally, we can add an LPIPS loss on the predicted clean signal when required. At sampling
time we run standard Euler/Heun on $\dot{\vx}=\vu$ with $\vu$ from
Eq.~(\ref{eq:mflow_recover_op}).

\subsection{Toy Example}
\label{sec:toy_example}

\begin{wrapfigure}{r}{0.49\textwidth}
    \centering
    \includegraphics[width=\linewidth]{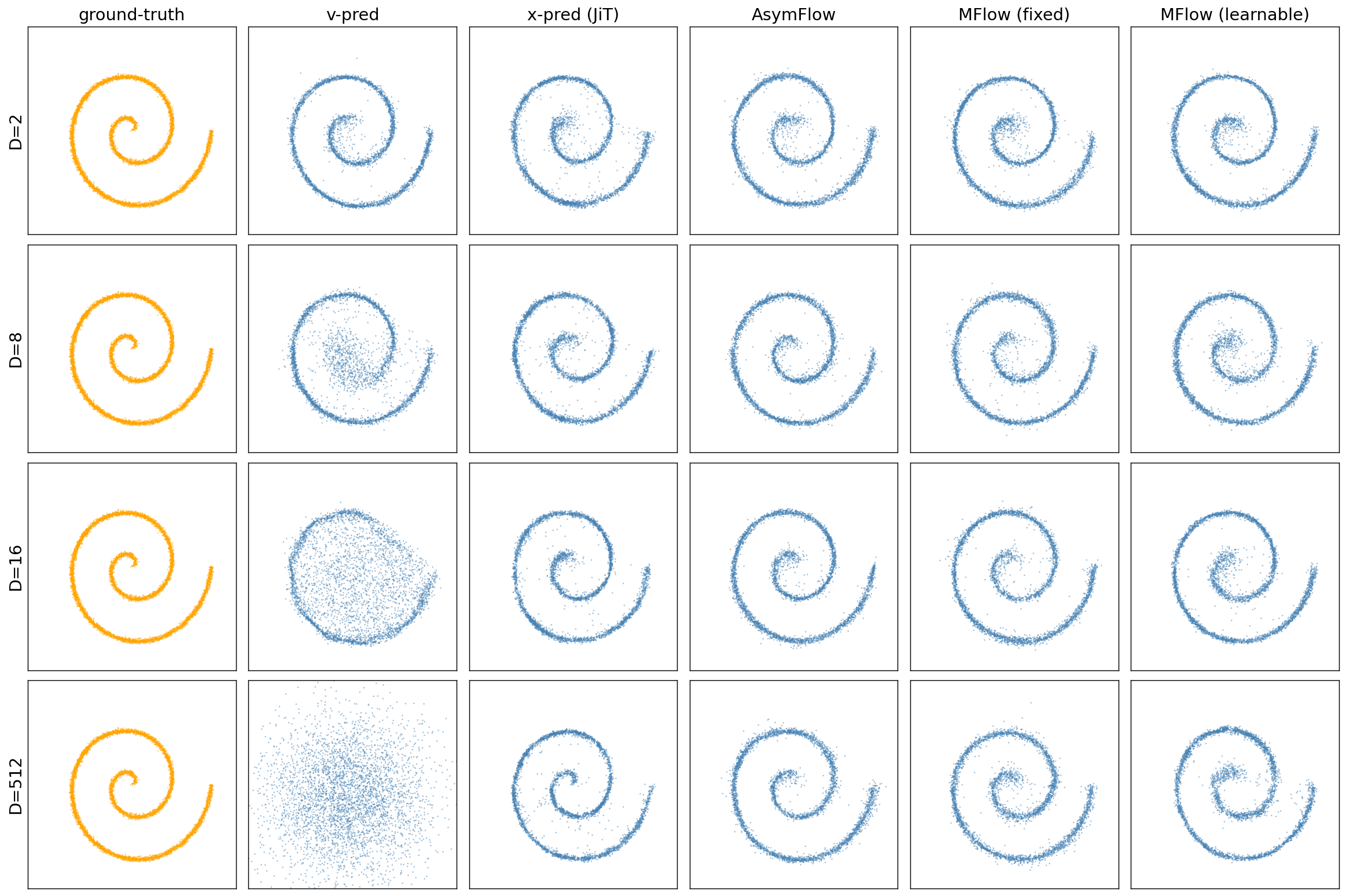}
    \vspace{-1.7em}
    \captionof{figure}{
    \textbf{Toy manifold experiment.}
    A 2-D spiral is embedded in
    $D\in\{2,8,16,512\}$ dimensions and projected back for visualization.
    Full $\vu$-prediction degrades with $D$, while $\vx_0$-prediction,
    rank-matched AsymFlow, and fixed or learned \method recover the spiral,
    including in the undercomplete $D=512$ setting.
    }
    \label{fig:toy_mflow}
\end{wrapfigure}

We revisit the toy manifold experiment of JiT (\cite{jit}) to test two
properties of \method: \textit{(a)} whether continuous, non-binary \textit{m}-values remain tractable
in high-dimensional observation spaces, and \textit{(b)} whether they can be
learned jointly with the generator. We sample a
2D spiral and embed it into a $D$-dimensional space using a
fixed random column-orthogonal projection. We fix the intrinsic dimension to $d=2$, vary
$D\in\{2,8,16,512\}$, and train a five-layer ReLU MLP with 256 hidden
units. We compare $\vx_0$-prediction, full $\vu$-prediction,
AsymFlow with $r=d$, \method with fixed
normalized-variance gains, and \method with gains learned from the same
initialization. All variants use the same architecture and
velocity-space training objective.
As shown in Fig.~\ref{fig:toy_mflow}, full $\vu$-prediction deteriorates
as $D$ increases, whereas $\vx_0$-prediction and rank-matched AsymFlow
remain stable. Both fixed and learnable \method also recover the spiral,
including at $D=512$. Thus,
high-dimensional tractability does not require binary gains
$m_i\in\{0,1\}$: continuous gains remain effective. The learnable variant further shows that this allocation can
be optimized end-to-end rather than specified in a heuristic manner.

\section{Signal Modeling}
\label{sec:sig_mod_in_jit}
\subsection{Motivation}
\label{sec:radial_freq_analysis}

\textbf{JiT lacks high-frequency but some other pixel diffusion models do not.} JiT's minimalist architecture lacks an explicit pathway dedicated to fine-grained signal details, like skip connections (\cite{uvit, hoogeboom2023simple, hoogeboom2025simpler}) or parallel refinement networks (\cite{dip, pixeldit}), motivating an investigation into whether this simplicity introduces limitations in its generated frequency distribution.
We study the frequency spectrum of the images generated by models we trained which share the JiT architecture. We use radially averaged power spectrum (RAPS) for this analysis. For context, images on a pixel grid can have at most $0.5$ cycles/pixel horizontal or vertical frequencies ($f_x$,$f_y$). However, in the case of radial averaging, the radial frequencies $f_r$ can reach till the diagonal frequency of $0.707$ ($=\sqrt{0.5^2+0.5^2}$). Interestingly, as seen in Fig.~\ref{fig:raps_jit_family}, we observe that all these models lack radial frequencies in the range of $0.5-0.7$. Recent concurrent works (\cite{wait, fvloss, fu2026frozen}) have also observed this.
As seen in Fig.~\ref{fig:raps_other_family}, we tried investigating other LDMs and pixel diffusion backbones. We find that neither VAE-reconstruction/generations of DiT (\cite{dit}), REPA-E (\cite{leng2025repa}), nor the generations of PixelDiT (\cite{pixeldit}) (which as a pixel level pathway) suffer from this issue, indicating that it might be a JiT backbone specific problem, more explicitely the use of high patchidication (and unpatchification) of $16-32\times$ without any intra-patch local operation.

\begin{table*}[h!]
\begin{minipage}[t]{0.45\linewidth}
\vspace{-12em}
\begin{minipage}[t]{\linewidth}
    \centering
    \includegraphics[width=0.82\linewidth]{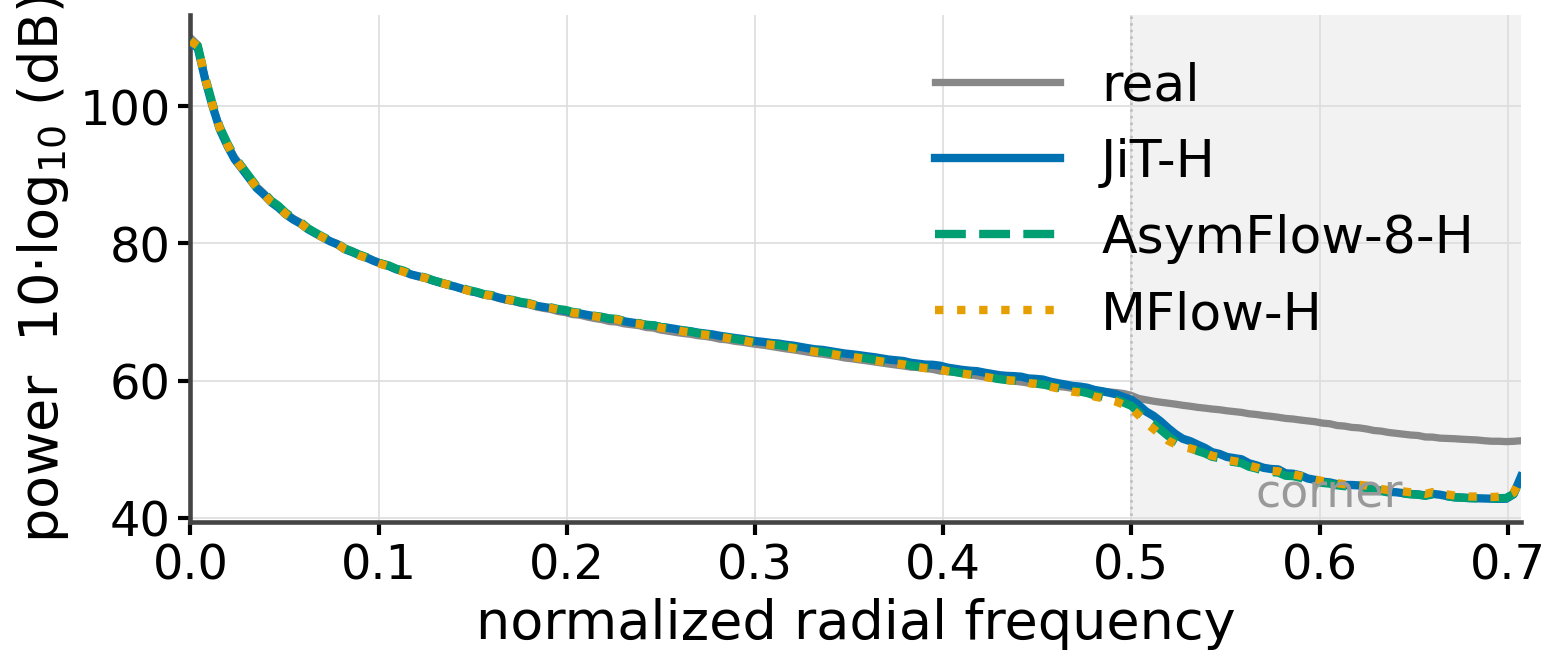}
    \vspace{-1em}
    \captionof{figure}{JiT-family generations lacks high frequency irrespective of parameterization. }
    \label{fig:raps_jit_family}
\end{minipage}
\vfill
\begin{minipage}[t]{\linewidth}
    \centering
    \includegraphics[width=0.82\linewidth]{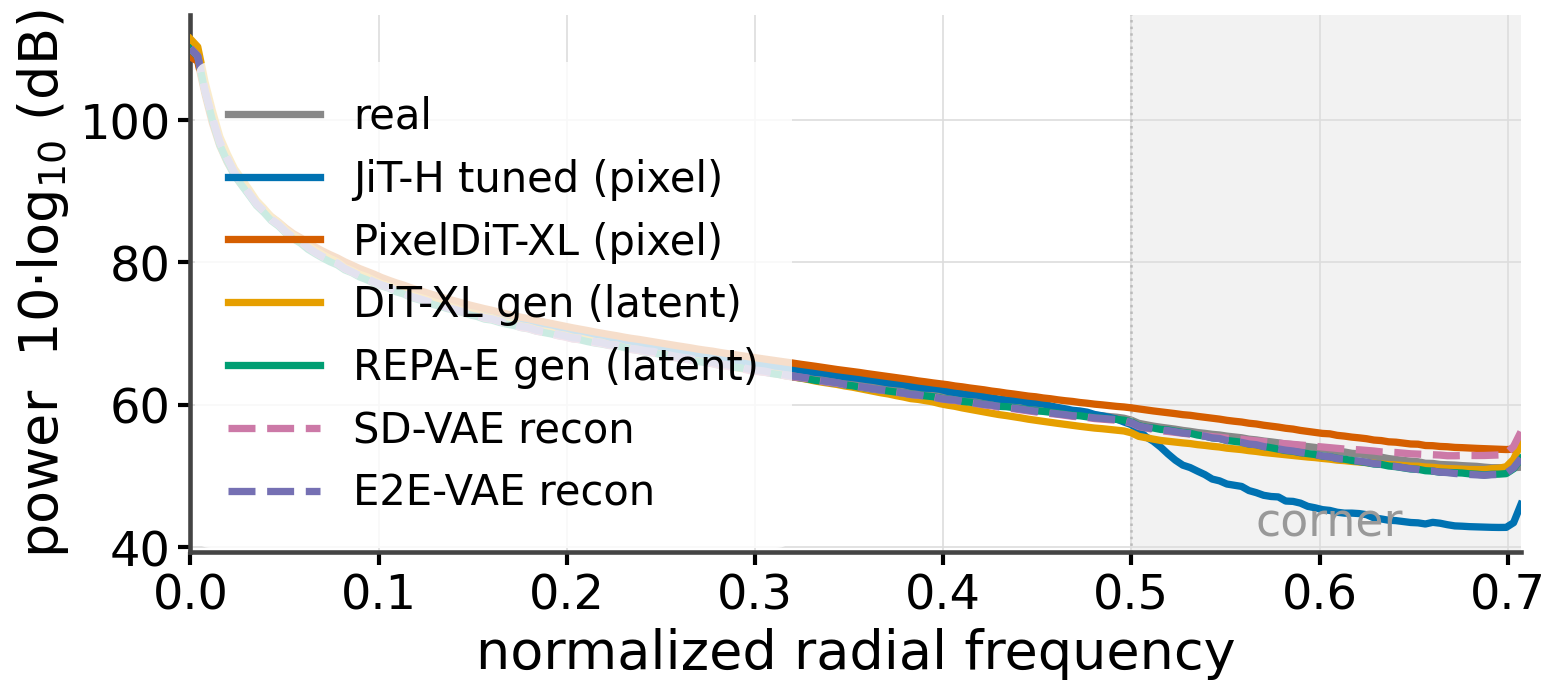}
    \vspace{-1em}
    \captionof{figure}{Latent model generations, VAE reconstructions, and other pixel diffusion models like PixelDiT do not lack high frequency.}
    \label{fig:raps_other_family}
\end{minipage}
\end{minipage}
\hfill
\begin{minipage}[t]{0.53\linewidth}
    \centering
    \includegraphics[width=\linewidth]{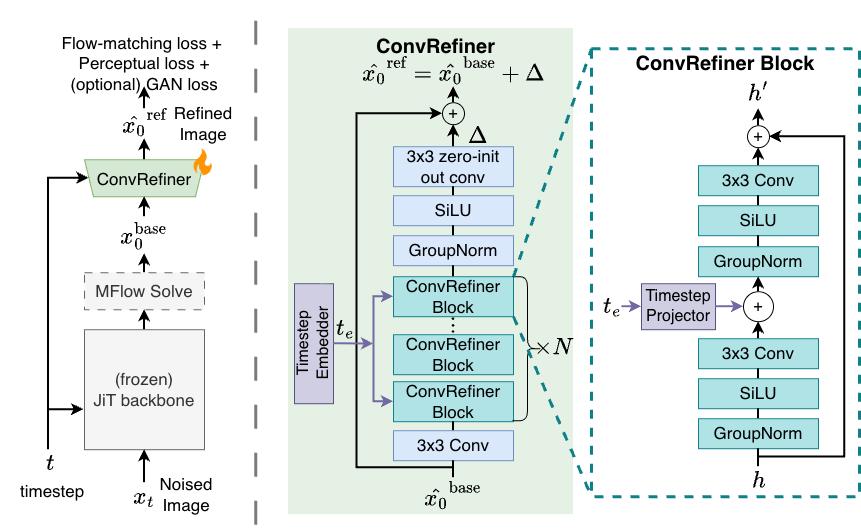}
    \captionof{figure}{\textbf{Overview of \emph{ConvRefiner} Formulation. }
    We propose a residual convolutional refiner post hoc training for a few epochs, with a fixed backbone, to regain the high-frequency similar to real images spectrum in vanilla vision transformers. 
    } 
    \label{fig:convrefiner_overview}
\end{minipage}
\end{table*}

\begin{wrapfigure}{r}{0.4\textwidth}
\vspace{-1em}
\begin{minipage}[t]{\linewidth}
    \centering
    \captionof{table}{\textbf{Summary of Results.} Our \method+ \emph{ConvRefiner} provides a flexible parameterization, saving expensive hyperparameter sweeps while having better FID, as well as solving the high-frequency deficit (measured in mean radial power deficit compared to real images) on ImageNet-1K 256$\times$256. }
    \label{tab:summary_in}
    \scalebox{0.57}{%
        \setlength{\tabcolsep}{0.2em}
        \begin{tabular}{lcc}
        \toprule
        \textbf{Method} & \textbf{FID\textdownarrow} & \textbf{HF-deficit}(dB)\\
        \midrule
        \cellcolor{gray!09}JiT-H/16 & \cellcolor{gray!09}\phantom{0}1.82 & \cellcolor{gray!09}-6.69\\
        \cellcolor{gray!20}\method-H/16 \method & \cellcolor{gray!20}\phantom{0}\textbf{1.81} & \cellcolor{gray!20}-6.63\\
        \cellcolor{gray!20}\method-H/16 + \emph{ConvRefiner} (ours) & \cellcolor{gray!20}\phantom{0}\textbf{1.81} & \cellcolor{gray!20}\textbf{-1.29}\\
        \bottomrule
        \end{tabular}
    }
\end{minipage}

\vspace{-1em}
\end{wrapfigure}

\subsection{ConvRefiner: A simple post-hoc fix}
Based on our analysis we suspect that this high-frequency deficit might be specific to the JiT-based pixel diffusion architecture. However, as seen in the previous section, convolutional VAE decoders are able to recover high-frequency.    \cite{ICLR2025_f062da19} note that only finetuning the last few layers of a pretrained VAE with an adversarial loss can improve local details at a reduced training cost.

Motivated by this, we add a few lightweight  convolutional blocks and post-hoc train only these blocks using an additional GAN loss apart from flow-matchting losses (Fig~\ref{fig:convrefiner_overview}). We keep the JiT backbone frozen to keep the refinement light. The \emph{ConvRefiner} predicts a residual over $\vx$-prediction of the network on which the perceptual and adversarial losses are then applied. For other parameterizations, the prediction is first converted to $\vx$ (clean image) and then the \emph{ConvRefiner} residual is applied on top of it.

\section{Experiments}

\vspace{-1em}
We extensively evaluate our proposed method on class-conditional ImageNet image generation to validate its effectiveness, as summarized in Table~\ref{tab:summary_in}. Then we delve into individual components -- \method and \emph{ConvRefiner}. We further show that \method can be extended to video generation and scaled to text-to-image generation. Finally, we ablate the design decision of each component.

\noindent\textbf{Datasets and Evaluation.}
For class-conditioned image generation experiments we train on ImageNet-1K (\cite{imagenet}). By default we show results on the standard 256${}\times{}$256 resolution images with the training and evaluation similar to \cite{jit}. For evaluation we use 
FID@50K (\cite{fid}).
We conduct class-conditioned video generation experiments on UCF-101 (\cite{ucf101}) dataset. We use 16 frames of 128${}\times{}$128 resolution videos for training and evaluation following \cite{wang2025larp}. Similar to images, 
FVD@10K (\cite{fvd})
serves as our video generation evaluation metric.
For text-to-image generation, we train on internal dataset of around 900M images (pretraining) and 33.5M (SFT), and evaluate on standard benchmarks of GenEval (\cite{ghosh2023geneval}), DPG-Bench (\cite{hu2024ella}), and PRISM-Bench (\cite{fang2026flux}). 

\begin{wrapfigure}{r}{0.4\textwidth}
\begin{minipage}[t]{\linewidth}
\vspace{-1em}
    \centering
    \captionof{table}{\textbf{Effect of LPIPS on ImageNet 256\texttimes 256 generation.} We see that LPIPS helps all the models in FID (evaluated on base JiT-B evaluation hyperparameters). However, it improves \method performance the most, beating the baselines.}
    \label{tab:imagnet_b}
    \scalebox{0.6}{%
        \setlength{\tabcolsep}{0.2em}
        \begin{tabular}{lcc}
        \toprule
        \multirow{2}{*}{\textbf{Method}} &  \multicolumn{2}{c}{\textbf{FID\textdownarrow}}  \\
        \cmidrule[\cmidrulewidth](l){2-3}
        & w/o LPIPS & w/ LPIPS\\
        \midrule
        \cellcolor{gray!09}JiT-B/16 (\cite{jit}) & \cellcolor{gray!09}\phantom{0}3.70 & \cellcolor{gray!09}3.42\\
        \cellcolor{gray!09}AsymFlow-B/16 (\cite{asym}) & \phantom{0}\cellcolor{gray!09}3.59 & \cellcolor{gray!09}3.40\\
        \cellcolor{gray!20}\method-B/16 (ours) & \cellcolor{gray!20}\phantom{0}3.74 & \cellcolor{gray!20}\textbf{3.33}\\
        \bottomrule
        \end{tabular}
    }
\end{minipage}
\end{wrapfigure}

\noindent\textbf{Implementation Details.} We build on top of the JiT repository and use their architecture for conditional image and video generation experiments. Hence, we use JiT both to refer to the vanilla transformer backbone and \xpred-prediction parameterization (especially when comparing against \method parameterization). We refer the base, large, and huge sizes as B-/L-/H-level. For these experiments, for ImageNet we used patch size of $16\times16$ leading to 256 tokens (with additional 32 in-context tokens like JiT), while for video we use $4\times8\times8$ patch size (with 4 as the temporal patch size) leading to 1024 tokens. For text-to-image, we train with a patch size of $16\times16$ on mixed-resolution 256p and 512p images with LPIPS loss ($0.1$ loss weighting).
For calculating patch-wise PCA-basis we follow AsymFlow. 
In the case of \emph{ConvRefiner}, we use a PatchGAN (\cite{isola2017image}) to apply the GAN loss only on the near-clean images (see appendix for details).

\begin{table*}[t!]
\begin{minipage}[t]{0.48\linewidth}
\centering
    \caption{\textbf{ImageNet 256\texttimes 256 comparisons}. We trained JiT, AsymFlow, and \method from scratch with additional LPIPS (\cite{lpips}) loss (but without REPA (\cite{yu2025repa})) and evaluated under the JiT protocol with individual hyperparameter tuning (denoted by *). We use the best hyperparameter for each after doing a sweep. The AsymFlow number remains in the range of the non-REPA performance noted in \cite{asym}. We use AsymFlow's FLOP convention as reported in \cite{pixeldit}.}
    \label{tab:imagenet_compare}
    \vspace{1em}
    \scalebox{0.58}{%
        \setlength{\tabcolsep}{0.2em}
        \begin{tabular}{lcccl}
        \toprule
        \textbf{Method} & \textbf{Pred (\textpm)} & \textbf{Params} & \textbf{GFLOPs} & \textbf{FID\textdownarrow} \\
        \midrule
        \multicolumn{5}{l}{\textit{Hierarchical CNNs (skip connections / U-Net-like (\cite{unet})}} \\
        \midrule
        ADM-G (\cite{adm}) & $\bm{\epsilon}$ & 554M & 2240 & 4.59 \\
        \midrule
        \multicolumn{5}{l}{\textit{Hierarchical transformers (skip connections / U-ViT-like (\cite{uvit}))}} \\
        \midrule
        RIN (\cite{rin}) & $\bm{\epsilon}$ & 320M & 668 & 3.42 \\
        SiD, U-ViT/2 (\cite{hoogeboom2023simple}) & $\bm{\epsilon}$ & 2B & 1110 & 2.44 \\
        VDM++, U-ViT/2 (\cite{vdmpp}) & $\bm{\epsilon}$ & 2B & 1110 & 2.12 \\
        SiD2, U-ViT/1 (\cite{hoogeboom2025simpler}) & $\bm{\epsilon}$ & - & 1306 & \textbf{1.38} \\
        \midrule
        \multicolumn{5}{l}{\textit{Hierarchical transformers (decoder head / DDT-like (\cite{ddt}))}} \\
        \midrule
        DiP-XL/16 (\cite{dip}) & $\vx_0-\bm{\epsilon}$ & 631M & - & 1.79 \\
        DeCo-XL/16 (\cite{DeCo}) & $\vx_0-\bm{\epsilon}$ & 682M & 245 & \underline{1.62} \\
        PixelDiT-XL/16 (\cite{pixeldit}) & $\vx_0-\bm{\epsilon}$ & 797M & 311 & \textbf{1.61} \\
        \midrule
        \multicolumn{5}{l}{\textit{Vanilla transformers (DiT-like  (\cite{dit}))}} \\     
        \midrule
        PixelFlow-XL/4 (\cite{chen2025pixelflow}) & $\vx_0-\bm{\epsilon}$ & 677M & 5818 & 1.98  \\
        PixelGen-XL/16 (\cite{pixelgen}) & $\vx_0$ & 676M & 260 & 1.83 \\
        PixelREPA-H/16 (\cite{pixelrepa}) & $\vx_0$ & 953M & 363 & \underline{1.81} \\
        \cellcolor{gray!09}JiT-H/16 (\cite{jit}) & \cellcolor{gray!09}$\vx_0$ & \cellcolor{gray!09}953M & \cellcolor{gray!09}363 & \cellcolor{gray!09}1.82* \\
        \cellcolor{gray!09}AsymFlow-H/16 (\cite{asym}) & \cellcolor{gray!09}$\vx_0-\mP \bm{\epsilon}$ & \cellcolor{gray!09}953M & \cellcolor{gray!09}363 & \cellcolor{gray!09}\textbf{1.79}* \\
        \cellcolor{gray!20} \method-H/16 (Ours) & \cellcolor{gray!20}$\vx_0-\mM \bm{\epsilon}$ & \cellcolor{gray!20}953M & \cellcolor{gray!20}363 & \cellcolor{gray!20}\underline{1.81}* \\
        \bottomrule
        \end{tabular}
    }
\end{minipage}
\hfill
\begin{minipage}[t]{0.48\linewidth}
\begin{minipage}[t]{\linewidth}
    \centering
    \caption{\textbf{UCF-101 video comparisons (B-level).} We show results on models we trained with best-tuned hyperparameters for each.}
    \label{tab:ucf_b}
    \scalebox{0.5}{%
        \setlength{\tabcolsep}{0.2em}
        \begin{tabular}{lc}
        \toprule
        \textbf{Method} & \textbf{FVD\textdownarrow}  \\
        \midrule
        \multicolumn{2}{l}{\textit{Latent diffusion models}} \\
        \midrule
        \cellcolor{gray!09}WAN-VAE2.1-1.3B(4×8×8)  + latent-DiT-B/1×1×1 & \cellcolor{gray!09}\phantom{0}169  \\
        \midrule
        \multicolumn{2}{l}{\textit{Pixel diffusion models}} \\
        \midrule
        \cellcolor{gray!09}Video JiT-B/4×8×8 (\cite{jit}) & \cellcolor{gray!09}\phantom{0}152  \\
        \cellcolor{gray!09}Video AsymFlow-B/4×8×8 (\cite{jit}) & \cellcolor{gray!09}\phantom{0}126  \\
        \cellcolor{gray!20}Video \method-B/4×8×8 (ours) & \cellcolor{gray!20}\phantom{0}\textbf{122}  \\
        \bottomrule
        \end{tabular}
    }
\end{minipage}
\vfill
\begin{minipage}[t]{\linewidth}
    \centering
    \vspace{1em}
    \caption{\textbf{UCF-101 video comparisons (L-level).} The rows in gray were trained by us and evaluated with individual hyperparameter tuning. $^\dagger$Note that we use the mapping network as generator and synthesis network as tokenizer convention from \cite{ren2026nerv}.}
    \label{tab:ucf_l}
    \scalebox{0.48}{%
        \setlength{\tabcolsep}{0.2em}
        \begin{tabular}{lcccc}
        \toprule
        \multirow{2}{*}{\textbf{Method}} & \multicolumn{2}{c}{\textbf{Params}} & \multirow{2}{*}{\textbf{\#Tokens}} & \multirow{2}{*}{\textbf{FVD\textdownarrow}}  \\
        \cmidrule[\cmidrulewidth](l){2-3} \\
        & \;Tokenizer\; & \; Generator\; & &\\
        \midrule
        \multicolumn{4}{l}{\textit{Autoregressive Models}} \\
        \midrule
        CogVideo (\cite{hong2022cogvideo}) & -- & 9.4B &2065& \phantom{0}625 \\
        TATS (\cite{ge2022long}) & 32M & 321M & 1024& \phantom{0}332 \\
        MAGVIT-AR (\cite{yu2023magvit}) & 158M & 306M & 1024& \phantom{0}265 \\
        OmniTokenizer (\cite{wang2024omnitokenizer}) & 82.2M & 650M & 1280& \phantom{0}191 \\
        MAGVITv2-AR (\cite{yu2024language}) & -- & 840M & 1280& \phantom{0}109 \\
        LARP (\cite{wang2025larp}) & 173M & 343M & 1024& \phantom{0}102 \\
        \midrule
        \multicolumn{4}{l}{\textit{Implicit weight-space generation}} \\
        \midrule
        DIGAN (\cite{yu2022generating})  & 58M$^\dagger$ & 5.5M$^\dagger$ & -- & \phantom{0}465 \\
        NeRV-Diffusion-L (\cite{ren2026nerv}) & $\sim$229M & 467M & 128& \phantom{0}\underline{97} \\
        \midrule
        \multicolumn{4}{l}{\textit{Pixel diffusion models}} \\
        \midrule
        VideoFusion (\cite{luo2023videofusion}) & -- & 2B & -- & \phantom{0}173 \\
        \cellcolor{gray!10}Video JiT-L/4×8×8 (\cite{jit}) &\cellcolor{gray!10}-- & \cellcolor{gray!10}458M& \cellcolor{gray!10}1024& \cellcolor{gray!10}\phantom{0}111 \\
        \cellcolor{gray!10}Video AsymFlow-L/4×8×8 (\cite{asym}) &\cellcolor{gray!10}-- & \cellcolor{gray!10}458M& \cellcolor{gray!10}1024& \cellcolor{gray!10}\phantom{0}\phantom{0}98 \\
        \cellcolor{gray!20}Video \method-L/4×8×8 (ours) &\cellcolor{gray!20}-- & \cellcolor{gray!20}458M& \cellcolor{gray!20}1024& \cellcolor{gray!20}\phantom{0}\phantom{0}\underline{97} \\
        \cellcolor{gray!20}\phantom{0}\phantom{0}\phantom{0}+ post-hoc Framewise-LPIPS (ours) &\cellcolor{gray!20}-- & \cellcolor{gray!20}458M&  \cellcolor{gray!20}1024 &\cellcolor{gray!20}\phantom{0}\phantom{0}\textbf{94}  \\
        \bottomrule
        \end{tabular}
    }
\end{minipage}
\end{minipage}
\end{table*}

\noindent\textbf{ImageNet.} \emph{(a) \method:} On class-conditioned image generation on
ImageNet 256\texttimes 256 experiment at B-level, as shown in Table~\ref{tab:imagnet_b}, without LPIPS loss and at the base JiT-B parameters, we observe that AsymFlow works better than both JiT and \method. Based on PixelGen (\cite{pixelgen}) we tried to add LPIPS loss to our baselines and \method model, and found that it provides performance gain across the board. Surprisingly, we see that this addition makes \method the best at this configuration. Subsequently, we train the same models with LPIPS at H-scale as shown in Table~\ref{tab:imagenet_compare}. We hypothesize the very slight lead of AsymFlow as a result of it being heavily tuned for ImageNet. Even so, we find that \method is very competitive to all the prior pixel diffusion transformer models without any parameterization search. 
\emph{(b) ConvRefiner:} We experiment with training the JiT and \method models on ImageNet $256\times256$ using our \emph{ConvRefiner} for an additional 30 epochs on top of the frozen 600 epoch trained base model. As shown in Fig.~\ref{fig:raps_fixed} (left), at both B- and H-levels and across both JiT and \method we see that the high radial frequency deficit is fixed. At the same time,  Fig.~\ref{fig:raps_fixed} (right) shows that this training maintains the FID. Hence, \emph{ConvRefiner} is effective at solving the high-frequency problem without affecting the FID at all. We provide some visual improvement results in Fig.~\ref{fig:convrefiner_qual}.

\begin{table*}[t]
\begin{minipage}[t]{0.48\linewidth}
    \centering
    \includegraphics[width=1\linewidth]{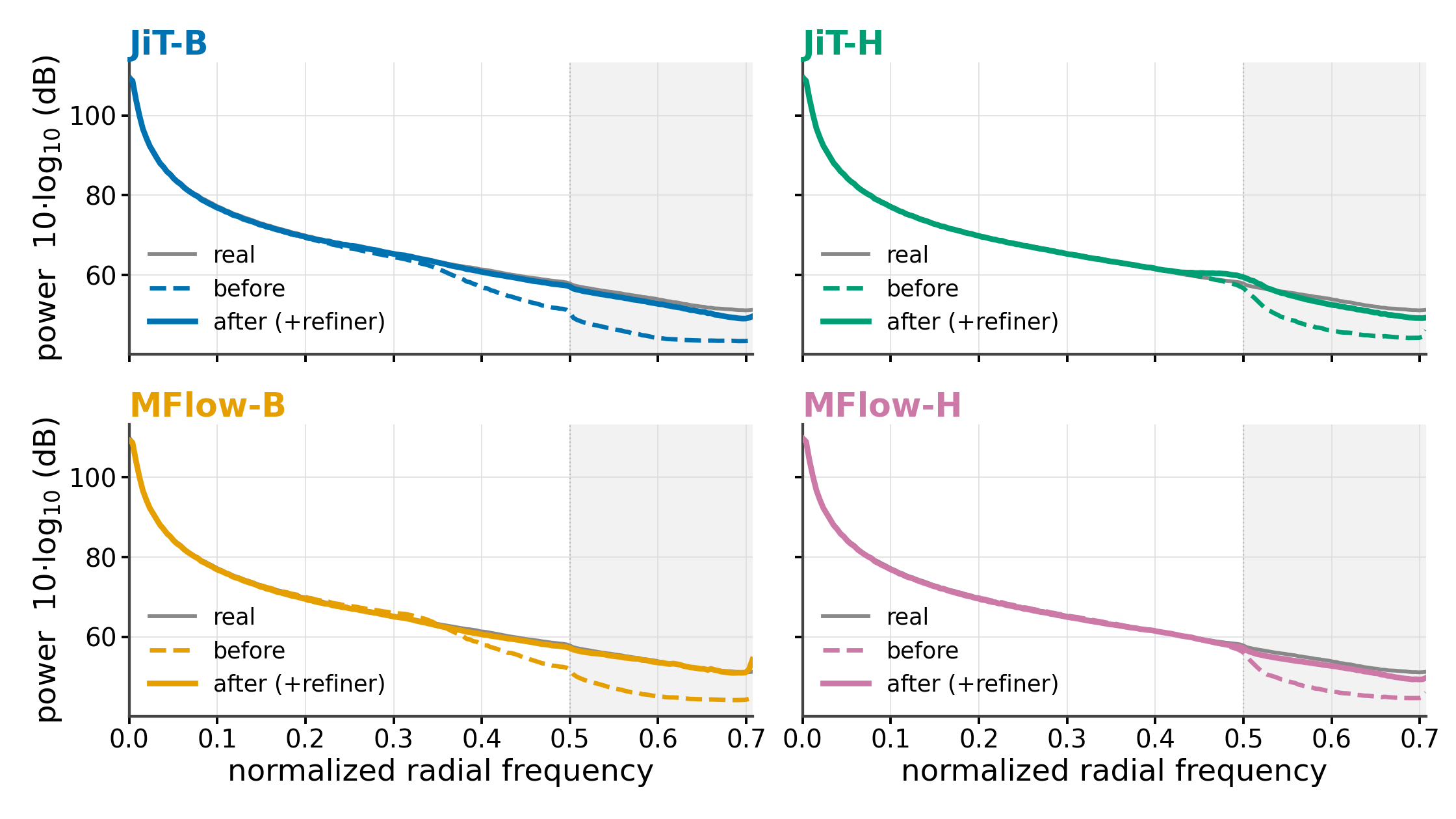}
\end{minipage}
\hfill
\begin{minipage}[t]{0.48\linewidth}
    \centering
    \includegraphics[width=\linewidth]{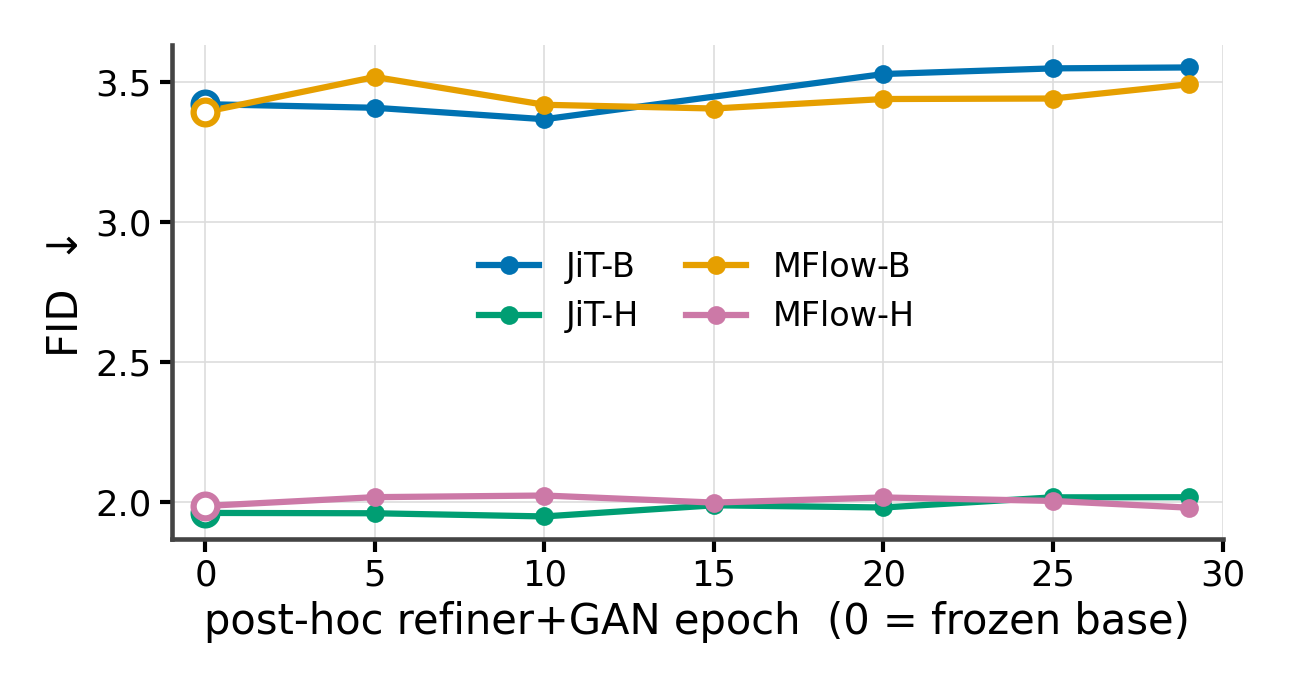}
\end{minipage}
\captionof{figure}{(Left) \emph{ConvRefiner} recovers high frequency across JiT/\method and model sizes B/H. (Right) The generative performance remains almost the same in this post-training.}
\label{fig:raps_fixed}
\vspace{-1em}
\end{table*}

\begin{figure}[h]
\begin{center}
\includegraphics[width=0.98\linewidth]{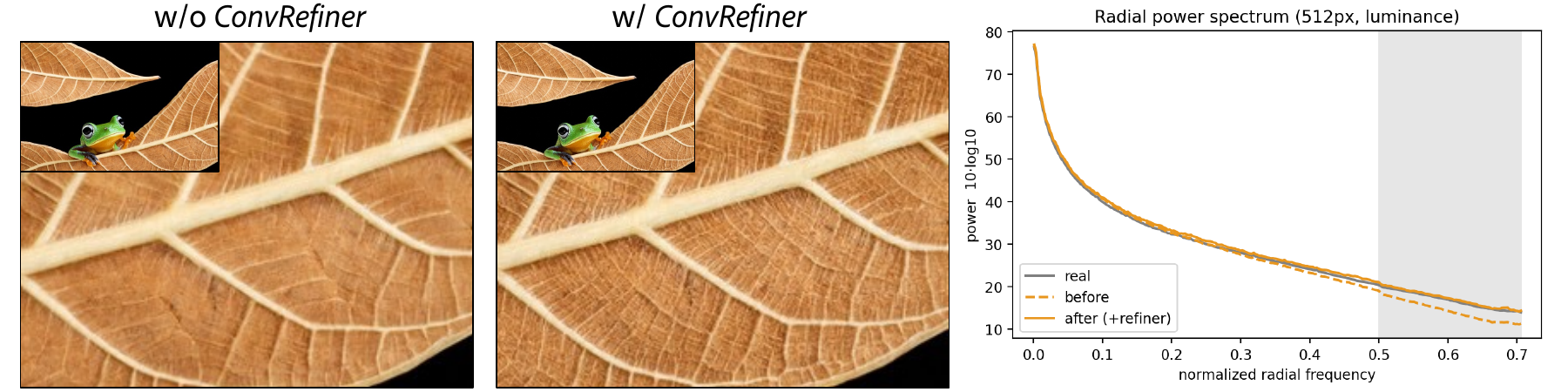}
\end{center}
\caption{\textbf{Visualizing effect of \emph{ConvRefiner} for text-to-image generation.} Here we visualize the effect of adding our post-training to \method-XXL on text-to-image generations. \textbf{(left)} We see improved details in the leaf veins in the generated image. \textbf{(right)} We also note the average power gain in the high-frequency (0.5–0.7) region. We see overall improvement in texture details comparable to real image spectrum.}
\label{fig:convrefiner_t2i}
\end{figure}

\noindent\textbf{Videos.}
We show the effectiveness of \method in the domain of videos apart from images. At the B-level class-conditioned UCF-101 video generation as shown in Table~\ref{tab:ucf_b}, we see that \method outperforms the WAN2.1 (\cite{wan2.1}) latent diffusion model, pixel-based JiT, and AsymFlow models we trained. 
We expand to L-level, where \method beats JiT and AsymFlow. With a small post-hoc Framewise-LPIPS training, we achieve best performance. Notably, we find training with Framewise-LPIPS from start (like ImageNet) hurts video generation.

\noindent\textbf{Text-to-Image.}
\emph{(a) \method:} We study whether \method scales beyond class-conditioned generation to
large-scale text-to-image (T2I). We instantiate JiT, AsymFlow
($r=8$), and \method using the same 1.1B-parameter MMDiT-L/16 backbone and
compare them under the same SFT setting. This controlled comparison isolates the effect of prediction parameterization while keeping the backbone and reported model size fixed. As shown in Table~\ref{tab:t2i_benchmark_comparison}, the three
parameterizations perform similarly at this scale, but \method is best
on every benchmark, even when compared against 
previous pixel-space T2I models.
Figure~\ref{fig:t2i-training} presents the benchmark progression over
training iterations. 
We further scale \method to an MMDiT-XXL/16 mixture-of-experts (MoE) model with
13B total parameters and 2.9B active parameters that
achieves the best performance among all models in the
comparison. 
Interestingly, model scaling also changes the parameterization learned by
\method as seen in Sec.~\ref{sec:discussion}, and \method can leverage this capacity automatically without any manual tuning.
\emph{(b) ConvRefiner:} 
We further apply \emph{ConvRefiner} as a post-training stage for text-to-image generation with \method-XXL. Starting from the SFT checkpoint, we train the \emph{ConvRefiner} for an additional 5k steps on top of the frozen base model. As shown in Fig.~\ref{fig:convrefiner_t2i} (right), without \emph{ConvRefiner} we observe the same high-frequency deficit in the $0.5$--$0.7$ band, even though it seems to be somewhat alleviated with increased model capacity compared to Fig.~\ref{fig:raps_fixed} (left). \emph{ConvRefiner} completely closes in this gap. This also shows up visually: Fig.~\ref{fig:convrefiner_t2i} (left) shows sharper fine-scale texture, such as the veins of the leaf. Hence, \emph{ConvRefiner} also fixes the high-frequency problem at scale for text-to-image generation. Additionally, as seen in \ref{tab:t2i_benchmark_comparison}, \emph{ConvRefiner} does not have noticable effect on text-to-image benchmark performance.

\begin{table*}[t]
    \centering

    \begin{minipage}[t]{0.29\textwidth}
        \vspace{0pt}
        \centering

        \includegraphics[width=0.9\linewidth]{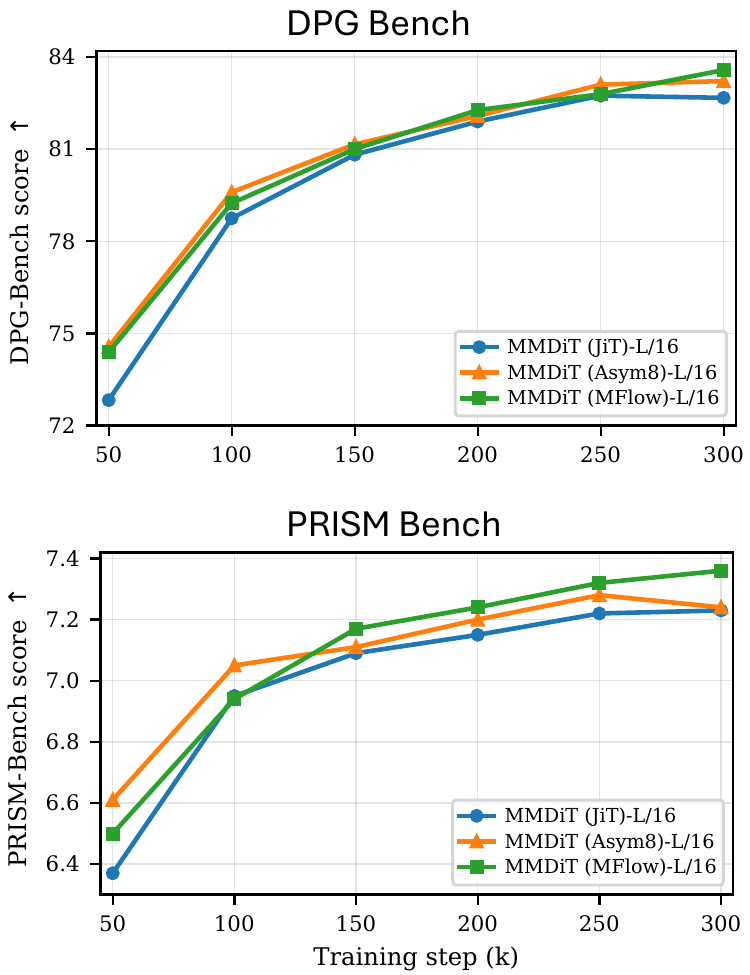}
        \vspace{-8pt}
        \captionof{figure}{Text-to-Image scores over training steps.}
        \label{fig:t2i-training}

    \end{minipage}
    \hfill
    \begin{minipage}[t]{0.69\textwidth}
        \vspace{0pt}
        \centering

        \caption{
            Comparison on text-to-image generation benchmarks.
            Parameter counts include only the image-generation model and
            exclude the text encoder. Higher is better for all metrics.
            Best results are in \textbf{bold}, and second-best results are
            \underline{underlined}.
        }
        \label{tab:t2i_benchmark_comparison}

        \vspace{-0.45em}

        \resizebox{\linewidth}{!}{%
            \renewcommand{\arraystretch}{1.08}
            \setlength{\tabcolsep}{3.2pt}

            \begin{tabular}{@{}lcccc@{}}
                \toprule
                \textbf{Model}
                & \textbf{Params}
                & \textbf{GenEval} $\uparrow$
                & \textbf{DPG} $\uparrow$
                & \textbf{PRISM} $\uparrow$ \\
                \midrule

                PixelDiT-T2I (\cite{pixeldit})
                & 1.3B
                & 0.7359
                & 84.58
                & 69.6 \\

                PixNerd-XXL/16 (\cite{wang2026pixnerd})
                & 1.2B
                & 0.7015
                & 81.61
                & 65.7 \\

                PixelFlow-XL/4 (\cite{chen2025pixelflow})
                & 882M
                & 0.6463
                & 77.90
                & 67.6 \\

                PixelGen-XXL/16 (\cite{pixelgen})
                & 1.1B
                & 0.7835
                & 79.62
                & 64.7 \\

                DeCo-XXL/16 (\cite{DeCo})
                & 1.1B
                & 0.8600
                & 81.96
                & 66.5 \\

                MiniT2I-L/16 (\cite{minit2i2026})
                & 912M
                & 0.8822
                & 84.66
                & 72.9 \\

                \midrule

                \rowcolor{gray!9}
                MMDiT (JiT)-L/16-SFT*
                & 1.1B
                & 0.8839
                & 84.65
                & 73.6 \\

                \rowcolor{gray!9}
                MMDiT (Asym8)-L/16-SFT*
                & 1.1B
                & 0.8832
                & 84.61
                & 73.9 \\

                \rowcolor{gray!20}
                MMDiT (\method)-L/16-SFT* (Ours)
                & 1.1B
                & 0.8843
                & 84.70
                & 73.9 \\

                \midrule

                \rowcolor{gray!20}
                MMDiT (\method)-XXL/16-SFT* (Ours)
                & 13B-MoE (2.9B active)
                & \textbf{0.8898}
                & \underline{86.10}
                & \underline{75.2} \\

                \rowcolor{gray!20}
                \phantom{0}\phantom{0}\phantom{0}\phantom{0}\phantom{0}\phantom{0}\phantom{0}\phantom{0}\phantom{0}\phantom{0}\phantom{0}\phantom{0}\phantom{0}\phantom{0}\phantom{0}\phantom{0}
                + \emph{ConvRefiner} (Ours)
                & 13B-MoE (2.9B active)
                & \underline{0.8896}
                & \textbf{86.16}
                & \textbf{75.4} \\

                \bottomrule
            \end{tabular}%
        }
    \end{minipage}
\end{table*}

\noindent\textbf{\method Ablations.} We ablated multiple axes of choices for \method on ImageNet B-level model. We clamp the initializations to $[0.05, 0.95]$, as otherwise the $\mathrm{sigmoid}$ function will blow up $\mM$.  In terms of loss parameterization we see that using $\vu_\mathrm{G}$ loss over $\vu$ loss significantly deteriorates performance, bringing FID from
$3.66$ to $4.88$. Training on $\vu_\mathrm{G}$ creates
a trivial incentive to reduce $m_i$ toward $\vx_0$-prediction without
improving the recovered transport field
(Appendix~\ref{app:mflow_velocity_supervision}). Subsequently, by varying the $\mM$ initialization to the normalized-variance (`norm\_var') as described in \ref{sec:mflow_formulation}, we see slight improvement over 'asym8' initialization, FID reduces from $3.84$ to $3.80$. Given that `norm\_var' has a smoother dependence on the PCA direction's variance, we choose it as default. We also experiment with learning-rate of $\mM$ parameters, and see that having a higher learning rate multiplier seems to help a bit bringing FID down from $3.80$ to $3.74$. 
We use this final setting (\emph{init `norm\_var', $\vu$-loss, learnable $\mM$ w/ 10$\times{}$ lr-multiplier}) as our default unless specified otherwise.

\begin{figure}[h]
\begin{center}
\includegraphics[width=0.98\linewidth]{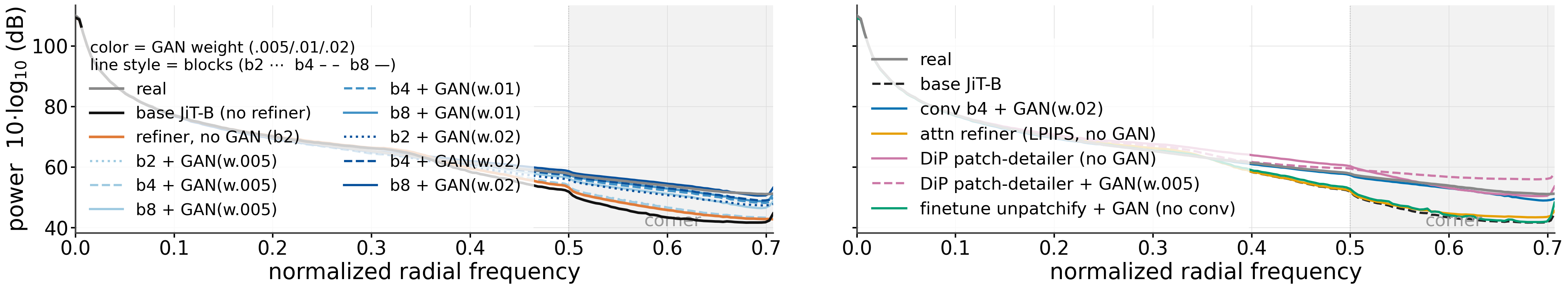}
\end{center}
\caption{(Left) Increasing blocks of Conv-refiner and the weight of GAN loss helps recover high frequency. (Right) Other baselines refining does not help except DiP patch detailer(\cite{dip}). This probably reflects a convolutions bias.}
\label{fig:raps_ablation}
\end{figure}

\noindent\textbf{\emph{ConvRefiner} Ablations.}
Our preliminary control experiment showed that JiT trained normally for 5\% more epochs does not improve the frequency spectrum. In Fig.~\ref{fig:raps_ablation} (left) we investigate the effect of number of blocks ($b\in{2,4,8}$) and the weight of the adversarial loss ($w\in{0.005,0.01,0.02}$). 
Firstly, we see that only using the refiner without the adversarial loss helps a bit.
However, the frequency spectrum gets closer to real with more blocks and more GAN weight. Favoring a smaller lightweight network, we finally select $b=4$ and $w=0.02$ as our final setting. This improves pixel level details as seen in Fig.~\ref{fig:convrefiner_ablation_qual}. 
Ablating the choice of refiner family (Fig.~\ref{fig:raps_ablation} (right)), we see that just finetuning the unpatchify layer or using a residual attention refiner has negligible effect. Another readily available convolutional refiner choice DiP patch-detailer(\cite{dip}) does improve the high-frequency and even overshoots with the GAN loss. However, it degrades the base JiT's FID 3.42 to 3.90 and 4.46 respectively, bolstering our choice of \emph{ConvRefiner}.

\section{Discussion}
\label{sec:discussion}
\begin{figure}[h]
\begin{center}
\includegraphics[width=\linewidth]{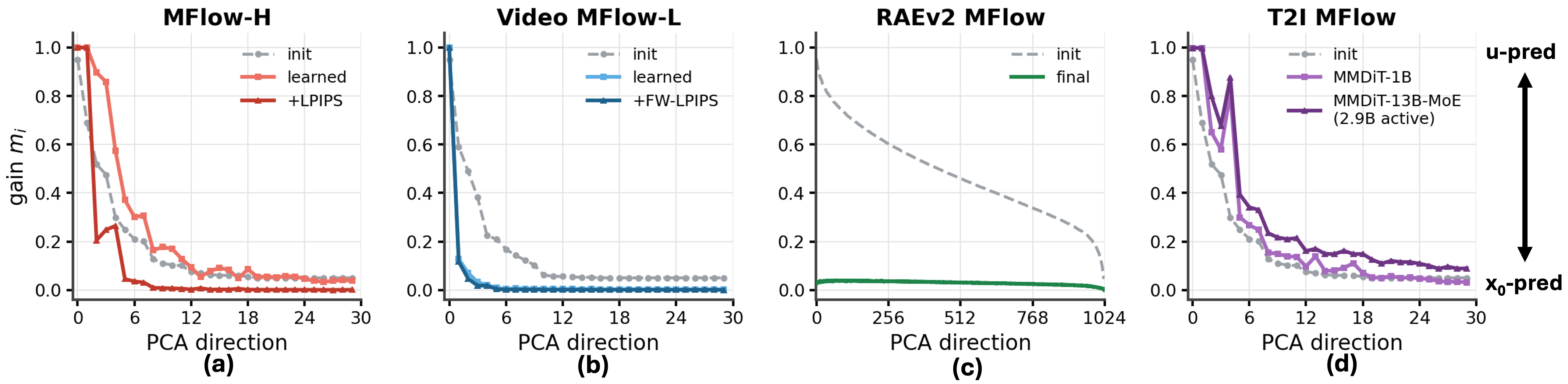}
\end{center}
\caption{\textbf{Learned $\mM$ values.} Variations in the \method gain curves across multiple axes: objectives -- w/ vs w/o LPIPS, domains -- images vs videos, representations -- pixel vs latent $\vx_0$-prediction models, model capacities -- 1B vs 13B-MoE.}
\label{fig:mflow_gain_curve}
\end{figure}
\noindent\textbf{$\mM$-curve Analysis.} Given the learned nature of the gain parameters $m_i$, we analyze how they vary across objectives, domains, representations, and model capacities. In Fig.~\ref{fig:mflow_gain_curve}, we visualize the learned $m_i$ values ordered by decreasing PCA variance, which we call an $\mM$-curve. Here, $m_i=0$ corresponds to $\vx_0$-prediction and $m_i=1$ to $\vu$-prediction; the `norm\_var' initialization is shown in gray.
\begin{enumerate}
    \item On ImageNet (a), adding LPIPS drives all but the leading few directions toward $\vx_0$-prediction while also improving FID (Table~\ref{tab:imagnet_b}), \emph{suggesting less need for $\vu$-like prediction beyond the dominant PCA directions}. 
    \item Moving to videos (b), the number of directions with non-negligible gains shrinks further, \emph{showing that the preferred parameterization varies across domains}. We discuss in detail about the behaviour of LPIPS on $\mM$-curve in \ref{app:lpips_mcurve}. 
    \item With RAEv2 (\cite{singh2026improved}) (c), whose latent model uses $\vx_0$-prediction, the gains collapse almost entirely toward zero, recovering an approximately pure $\vx_0$ parameterization. Yet forcing $\vu$-prediction on a small leading subspace changes FID only marginally (Table~\ref{tab:imagenet_rae}), \emph{indicating that a low-rank $\vu$ subspace can still be tolerated}. 
    \item Finally, in text-to-image generation (d), increasing model capacity shifts more directions toward $\vu$-prediction. Together with the RAEv2 result, this suggests that \emph{model capacity influences how much $\vu$-prediction a model can accommodate, whereas the statistics of the data representation determine where that $\vu$-prediction is most useful}. 
\end{enumerate}
In particular, \textbf{for a fixed budget of $N$ $\vu$-predicted directions, the allocation should remain data-dependent, with the leading high-variance PCA directions providing a natural choice}. \method can exploit increased capacity or adapt to domain change, without requiring a manually chosen rank.

\section{Conclusion}
We revisit two design questions in vanilla pixel diffusion transformers: how their prediction parameterization should adapt to the data, and whether their minimalist architecture captures high-frequency. We introduce \method, a direction-adaptive parameterization that learns a continuous interpolation between clean-data and velocity prediction across principal data directions, avoiding the need for manual subspace selection, saving parameterization search cycles.
Our analysis of the learned M-curves further shows that the preferred parameterization changes systematically with both data statistics and model capacity.
Beyond parameterization, we identify a persistent high-frequency deficit in JiT-style models and address it with \emph{ConvRefiner}, a lightweight post-hoc convolutional module that restores high-frequency statistics while preserving generative quality. Together, these results suggest that both adaptive prediction targets and explicit treatment of fine-scale signal structure are important ingredients for scaling simple pixel-space diffusion transformers.

\subsection*{AI use statement}

In this work, we used AI tools only in supporting non-research roles under author supervision. Specifically, we used it to correct grammatical errors and improve the clarity of author-written text, and to assist with software tooling. In particular, we used it for plotting and visualization code used to produce some of the figures in this paper. We did not use AI tools for research ideations, for deriving results, for designing experiments, or for generating any scientific claims. We have reviewed all the AI-assisted text and figures for correctness.

\subsection*{Reproducibility statement}
We provide the \method's closed-form velocity-recovery rule in Sec.~\ref{sec:mflow_formulation} and Python-like pseudocode in Sec.~\ref{sec:mflow_impl}. The post-hoc \emph{ConvRefiner} is specified as an architecture figure in Fig.~\ref{fig:convrefiner_overview}, and its training objective is provided in Section~\ref{sec:sig_mod_in_jit} and Appendix~\ref{sec:refiner_more_details}. Full training and sampling configurations like optimizer, learning-rate schedule, batch size, EMA, noise schedule, and the guidance interval, guidance scale, and number of sampling steps used at evaluation, are tabulated per model in Appendix~\ref{sec:hparams}.  Radially-averaged power spectrum (RAPS) evaluation protocol are described in Appendix~\ref{sec:raps}. All reported metrics (FID, FVD, RAPS, GenEval, DPG, Prism) are computed with standard public implementations we cite.

\subsubsection*{Acknowledgments}
We would like to thank Harsha Walvekar, Vatsal Agarwal, Prateksha Udhayanan, Namitha Padmanabhan, and Matthew Gwilliam for their help shaping this paper.

\bibliography{iclr2027_conference}
\bibliographystyle{iclr2027_conference}

\appendix
\section{Appendix}

\subsection{Hyperparameters}
\label{sec:hparams}
We provide our training hyperparameters in Table~\ref{tab:config}. For the ImageNet and UCF101 experiments we do not specifically change any of the base architecture, unless specified otherwise. Furthermore, we also provide evaluation time hyperparameters for these experiments in Table~\ref{tab:eval_hparams}. Because the model parameterization changes over training for \method and hence it is not clear if the defaults work. We report inference steps, classifier-free guidance (CFG),  sampling velocity-recovery clamp $\sigma_{\min}$, and  CFG Interval. 

We use the default JiT-B hyperparameters (Heun 50 steps, CFG 2.9, CFG Interval $[0.10,1.0]$, $\sigma_{\text{min}}=0.04$) for initial ImageNet B-level experiments to isolate the effect of training objective.
 For H-scale results we first conducted a coarse per-model CFG scan at $\sigma_{\min}=0.04$, interval $[0.10,1]$, followed by a finer CFG$\times\sigma_{\min}$ grid on the strongest (LPIPS) variant. 

For UCF-101, we firstly observed that a smaller spatial patch size of $8\times8$ worked better than $16\times16$. For the Wan2.1 latent model we used the `Wan-AI/Wan2.1-T2V-1.3B-Diffusers' VAE and a DiT based on JiT's modern architecture.  For evaluation hyperparameters, we use a two-stage search because full-scale FVD evaluation is expensive. We first run a FVD@2K grid over CFG 
 and 
 $\sigma_{\min}$
with Heun/100 to find the basin. 
And then over a narrowed CFG$\times\sigma_{\min}$ grid around the coarse optimum we calculate the full-scale FVD@10K. On the baseline, we also found $100$ steps of Heun works the best around the basin, so we use that for all the experiments. The \method + post-hoc Framewise-LPIPS loss model was trained for an additional 10\% epochs. (We note that, the UCF101 results in \cite{HPDM} is evaluated at a different spatial resolution of 256 with temporal subsampling and without a publicly available codebase or checkpoint it is not comparable. Hence, we omit them.) 

We additionally port the three objectives to a large text-to-image (T2I) stack: a multimodal
DiT (MMDiT) over a weight-free PixelVAE latent ($16\times16$ pixel patchify $\to$ $768$-d
tokens) with a frozen Qwen3-VL-8B text encoder (hidden layers $\{10,20,-1\}$). We train a dense
$1.1$B model and a $13$B mixture-of-experts ($2.9$B active) on ${\sim}900$M images
(multi-resolution $256/512$) with captions, then supervised-fine-tune for
$4$K steps on a high-quality mixture $33.5$M aesthetic-filtered images. All variants share the same natural-image patch-PCA basis
(uncentered, $16\times16$) as the ImageNet models, and \method again uses rational gains
($c{=}0.05$) with a $10\times$ gain learning rate. Unlike the ImageNet-H and UCF settings, we do
\emph{not} run a per-model CFG$\times\sigma_{\min}$ search for T2I. We fix a single
shared sampler for all variants (DDIM, $50$ steps, CFG $3.0$; Table~\ref{tab:eval_hparams}), so
the T2I numbers reflect out-of-the-box behavior.

\begin{table}[h]
\centering
\caption{Configuration for our JiT / AsymFlow / \method models. Architecture and optimizer follow JiT-B/L/H (\cite{jit} Tab.~9) for image/video; the text-to-image (T2I) stack is a PixelVAE-latent MMDiT. We list the settings we update.}
\label{tab:config}
\small
\setlength{\tabcolsep}{5pt}
\begin{tabular}{@{}llll@{}}
\toprule
 & \textbf{Image} (ImageNet-256) & \textbf{Video} (UCF-101, $16{\times}128^2$) & \textbf{Text-to-Image} \\
\midrule
\multicolumn{4}{@{}l}{\emph{Architecture}} \\
\midrule
Backbone         & JiT-B/H (JiT Tab.~9)   & VideoJiT-B/L (= JiT depth/width) & MMDiT \\
Params           & 131M / 953M            & 131M / 458M                     & 1.1B / 13B-MoE 2.9B act  \\
Patch            & $16\times16$           & $4{\times}8{\times}8$           & $16\times16$ (PixelVAE) \\
Sequence length  & 256 ($+32$ in-context) & 1024                            & res-dep. ($1024$ @\,$512$px) \\
Bottleneck       & 128 (B), 256 (H)       & 256                             & -- (none) \\
Text encoder     & -- (class label)       & -- (class label)                & Qwen3-VL-8B (frozen) \\
Latent           & pixel                  & pixel                           & pixel \\
Dropout          & 0 (B), 0.2 (H)         & 0                               & 0 \\
\midrule
\multicolumn{4}{@{}l}{\emph{Training}} \\
\midrule
Optimizer        & AdamW, $\beta{=}(0.9,0.95)$ & same             & AdamW , $\beta{=}(0.95,0.95)$\\
Epochs / warmup  & 600 / 5                & 3000 / 5                        & $300$K$+$4K steps / $1$K \\
Effective batch  & 1024                   & 512 (B), 256 (L)                & $2048$$^\dagger$ \\
Learning rate    & $2\!\times\!10^{-4}$   & $1\!\times\!10^{-4}$/$5\!\times\!10^{-5}$ & $1\!\times\!10^{-4}$ \\
LR schedule      & constant               & constant                        & warmup-stable-decay \\
EMA decay        & \{0.9996,0.9998,0.9999\} & same                          & 0.9995 \\
Time sampler     & logit-normal $\mu{=}{-}0.8,\sigma{=}0.8$ & same          & resolution-shifted$^\dagger$ \\
Precision        & bf16, no grad-clip     & same                            & bf16, grad-clip $1.0$ \\
Class drop (CFG) & 0.1                    & 0.1                             & 0.05 (text drop) \\
\midrule
\multicolumn{4}{@{}l}{\emph{AsymFlow / \method}} \\
\midrule
AsymFlow rank    & $r{=}8$                & $r{=}8$                         & $r{=}8$ \\
\method gains init & $m{=}\hat\lambda/(\hat\lambda{+}c)$, $c{=}0.05$ & same & same \\
\method $\theta$ & learnable, init `norm\_var' & same                       & learnable, init `type' (clamp) \\
\method loss space & velocity ($u$)       & velocity ($u$)                  & velocity ($u$) \\
\method gain LR mult. & $\times 10$       & $\times 1$                      & $\times 10$ \\
PCA basis        & fixed patch-PCA, ImageNet & fixed tubelet-PCA, UCF-101   & fixed patch-PCA, natural img \\
                 & (uncentered, $16{\times}16$) & (uncentered, $4{\times}8{\times}8$) & (uncentered, $16{\times}16$) \\
\midrule
\multicolumn{4}{@{}l}{\emph{LPIPS}} \\
\midrule
Backbone / gate  & VGG, $t\!\ge\!0.3$     & VGG, $t\!\ge\!0.3$              & VGG, $t\!\ge\!0.3$ \\
Weight           & 0.1                    & 0.2                             & 0.1 \\
Spatial extent   & full frame             & random $H/2{\times}W/2$         & full frame \\
Frames           & --                     & all 16 (framewise)              & -- \\
\bottomrule
\end{tabular}
\end{table}

\begin{table}[h!]
\centering
\caption{Evaluation / sampling hyperparameters for our trained models (Tables~\ref{tab:imagnet_b},~\ref{tab:imagenet_compare}, ~\ref{tab:ucf_b}, ~\ref{tab:ucf_l}). Table~1 uses a single shared base JiT-B configuration while the rest are tuned per model.}
\label{tab:eval_hparams}
\small
\setlength{\tabcolsep}{5pt}
\begin{tabular}{@{}llccccr@{}}
\toprule
Table & Model & Solver & Steps & CFG & $\sigma_{\min}$ & CFG Interval \\
\midrule

\multirow{3}{*}{\shortstack[l]{Table~\ref{tab:imagenet_compare}\\(ImageNet, H, tuned)}}
 & JiT-H $+$LPIPS            & Heun & 50 & 2.50 & 0.04 & $[0.10, 1.0]$ \\
 & AsymFlow-H $+$LPIPS       & Heun & 50 & 2.20 & 0.04 & $[0.10, 1.0]$ \\
 & \method-H $+$LPIPS          & Heun & 50 & 2.35 & 0.04 & $[0.10, 1.0]$ \\
\midrule
\multirow{2}{*}{\shortstack[l]{Table~\ref{tab:ucf_b}\\(UCF, B, tuned)}}
 & Video JiT-B               & Heun & 100 & 3.0 & $10^{-5}$ & $[0, 1.0]$ \\
 & Video AsymFlow-B & Heun & 100 & 3.5 & 0.04 & $[0, 1.0]$ \\
 & Video \method-B            & Heun & 100 & 3.5 & 0.04 & $[0, 1.0]$ \\
\midrule
\multirow{3}{*}{\shortstack[l]{Table~\ref{tab:ucf_l}\\(UCF, L, tuned)}}
 & Video JiT-L              & Heun & 100 & 3.0 & $10^{-5}$ & $[0, 1.0]$ \\
 & Video AsymFlow-L           & Heun & 100 & 4.0 & 0.04 & $[0, 1.0]$ \\
 & Video \method-L           & Heun & 100 & 4.0 & 0.04 & $[0, 1.0]$ \\
 & \quad $+$post-hoc FW-LPIPS & Heun & 100 & 3.0 & 0.04 & $[0, 1.0]$ \\
 \midrule
\multirow{4}{*}{\shortstack[l]{Tables~\ref{tab:imagenet_compare}$^{\ast}$\\(T2I, shared)}}
 & MMDiT JiT-L $+$SFT        & DDIM & 50 & 3.0 & $10^{-5}$ & $[0, 1.0]$ \\
 & MMDiT AsymFlow-L $+$SFT   & DDIM & 50 & 3.0 & 0.04 & $[0, 1.0]$\\
 & \method-L $+$SFT            & DDIM & 50 & 3.0 & 0.04 & $[0, 1.0]$ \\
 & \method-XXL (MoE) $+$SFT    & DDIM & 50 & 3.0 & 0.04 & $[0, 1.0]$ \\
\bottomrule
\end{tabular}
\end{table}

\smallskip
{\footnotesize\raggedright
$^{\dagger}$\,Normalized time $\tau{=}t/T$ (with $t/T{=}\sigma(z),\ z\!\sim\!\mathcal{N}(0,1)$) is warped as
$\tau\mapsto \dfrac{s\,\tau}{1+(s-1)\,\tau}$, where $s{=}\min\!\big(\sqrt{N/N_{\mathrm{ref}}},\,s_{\max}\big)$,
$N$ is the latent token count ($HW\!\cdot\!\min(T,4)$; images $T{=}1$), $N_{\mathrm{ref}}{=}256$ ($16{\times}16$),
and $s_{\max}{=}6$ (SD3/Hunyuan-style resolution shift).\par}

\clearpage
\subsection{\method Implementation Details}
\label{sec:mflow_impl}

All operators are diagonal in the fixed patch-PCA basis $\mU$, so every step is a change of basis
(\texttt{img$\to$patch coeffs @ U}), a per-direction scalar op, and a change back (\texttt{@ U$^\top\to$ img}).
Below, \texttt{x} is a batch of patch-token tensors, \texttt{t$\in$[0,1]} is flow matching timestep, \texttt{U (D,D)}
and \texttt{lam (D,)} are the precomputed PCA basis/variances. These mirror the JiT codebase.

\begingroup
\definecolor{mflowKw}{rgb}{0.0,0.5,0.0}     %
\definecolor{mflowCm}{rgb}{0.6,0.0,0.0}     %
\definecolor{mflowStr}{rgb}{0.0,0.3,0.6}    %
\definecolor{mflowFn}{rgb}{0.5,0.1,0.6}     %
\lstset{language=Python,%
        basicstyle=\ttfamily\footnotesize,columns=fullflexible,keepspaces=true,%
        keywordstyle=\color{mflowKw}\bfseries,commentstyle=\color{mflowCm},%
        stringstyle=\color{mflowStr},%
        emph={setup_gains,gains,mflow_target,mflow_to_x0,mflow_to_velocity,%
              training_loss,sample,to_patch_coeffs,to_img,clamp_min,sigmoid,%
              mse,weight,randn_like,heun_step,schedule,net,exp,log,randn},%
        emphstyle=\color{mflowFn}\bfseries,%
        breaklines=true,breakatwhitespace=true,showstringspaces=false,%
        xleftmargin=1em,aboveskip=4pt,belowskip=4pt,
        literate=
            {Eq_mgrad}{{Using Eq.~\ref{eq:mgrad}}}1
            {Eq_mflow_target}{{Using Eq.~\ref{eq:mflow_target}}}1
            {Eq_mflow_comp}{{Using Eq.~\ref{eq:mflow_comp}}}1
            {Eq_mflow_sys}{{Using Eq.~\ref{eq:mflow_sys}}}1
            {Eq_mflow_recover_comp}{{Using Eq.~\ref{eq:mflow_recover_comp}}}1
            {Eq_mflow_recover_op}{{Using Eq.~\ref{eq:mflow_recover_op}}}1
            {Eq_mflow_loss}{{Using Eq.~\ref{eq:mflow_loss}}}1
        }

\noindent\rule{\linewidth}{0.4pt}
\begin{lstlisting}
# One-time setup: fixed gains from lambda, or learnable
# params. lam_n is the normalized variance (lam_max = 1).
def setup_gains(lam):
  lam_n = lam.clamp(min=0) / lam.clamp(min=0).max()
  m_fixed = lam_n / (lam_n + c)
  return m_fixed

# Learnable variant (default): phi (D,) logits + scalar
# m_i = sigmoid(phi_i).
def gains():                     
  if learnable:
    m = sigmoid(phi)     # phi is init with torch.logit(m_fixed)
  else:
    m = m_fixed
  return m
\end{lstlisting}
\noindent\rule{\linewidth}{0.4pt}
\begin{lstlisting}
# Graded target u_G = x0 - M eps (Eq_mflow_target), in the PCA basis.
def mflow_target(x0, eps, t):
  m  = gains()
  ce = to_patch_coeffs(eps) @ U        # eps in PCA coords
  Me = (m * ce) @ U.T                  # M eps = U diag(m) U^T eps
  return x0 - to_img(Me)               # u_G
\end{lstlisting}
\noindent\rule{\linewidth}{0.4pt}
\begin{lstlisting}
# Exact recovery of x0-hat from the output u_G (Eq_mflow_recover_comp, Eq_mflow_recover_op).
def mflow_to_x0(u_M, x_t, t):
  m   = gains(t)
  omt = clamp_min(1 - t, t_eps)        # (1 - t)
  cu  = to_patch_coeffs(u_M) @ U       # u_M coeffs
  cz  = to_patch_coeffs(x_t) @ U       # x_t coeffs
  denom = clamp_min(omt + m * t, t_eps)  # Delta_i = (1-t)+m_i t
  x0_c  = (omt * cu + m * cz) / denom  # x0-hat coeffs (eq. 11)
  return to_img(x0_c @ U.T)
\end{lstlisting}
\noindent\rule{\linewidth}{0.4pt}
\begin{lstlisting}
# Velocity for sampling (Eq_mflow_recover_op).
def mflow_to_velocity(u_M, x_t, t):
  x0_hat = mflow_to_x0(u_M, x_t, t)
  return (x0_hat - x_t) / clamp_min(1 - t, t_eps)
\end{lstlisting}
\noindent\rule{\linewidth}{0.4pt}
\begin{lstlisting}
# Training step: velocity-space flow-matching loss.
def training_loss(net, x0, t):
  eps = randn_like(x0)
  x_t = t * x0 + (1 - t) * eps         # forward interpolant
  u_M = net(x_t, t, cond)              # raw output ~= u_G
  u_pred = mflow_to_velocity(u_M, x_t, t)  # -> full velocity
  loss   = weight(t) * mse(u_pred, x0 - eps)  # Eq_mflow_loss
  return loss
\end{lstlisting}
\noindent\rule{\linewidth}{0.4pt}
\begin{lstlisting}
# Sampling: Euler/Heun on velocity, integrating t: 0 -> 1.
def sample(net, steps):
  x = randn(...)                       # t = 0, pure noise
  for t0, t1 in schedule(steps):
    u = mflow_to_velocity(net(x, t0, cond), x, t0)
    x = heun_step(x, u, t0, t1, net)
  return x
\end{lstlisting}
\noindent\rule{\linewidth}{0.4pt}
\endgroup

\subsection{Why \method Uses $u$-prediction}
\label{app:mflow_velocity_supervision}

\method predicts the graded target
\begin{equation}
    \vu_{\mathrm G}
    =
    \vx_0-\mM\bm\epsilon,
\end{equation}
where the direction-wise gains
$m_i=\operatorname{sigmoid}(\phi_i)$ are optimized jointly with the
network. A natural alternative would be to directly regress this target:
\begin{equation}
    \mathcal L_{\mathrm G}(\vtheta,\vphi)
    =
    \E\left[
        w(t)
        \left\|
            \hat{\vu}_{\mathrm G}
            -
            \vu_{\mathrm G}
        \right\|_2^2
    \right].
    \label{eq:direct_graded_loss}
\end{equation}
Here, $\vu_{\mathrm G}$ is not detached from the computation graph, so
this objective does propagate gradients to the gains. We show, however,
that it introduces an undesirable dependence on the scale of the chosen
parameterization.

Recall that
\begin{equation}
    \vx_t=t\vx_0+(1-t)\bm\epsilon,
    \qquad
    \vu=\vx_0-\bm\epsilon.
\end{equation}
In PCA direction $i$, the graded target can be rewritten as
\begin{equation}
    \tilde u_{\mathrm G,i}
    =
    (1-m_i)\tilde x_{t,i}
    +
    \Delta_i(t)\tilde u_i,
    \qquad
    \Delta_i(t)\coloneqq(1-t)+tm_i.
    \label{eq:graded_target_velocity_form}
\end{equation}
Consequently, the velocity recovered from a network prediction
$\hat{\tilde u}_{\mathrm G,i}$ is
\begin{equation}
    \hat{\tilde u}_i
    =
    \frac{
        \hat{\tilde u}_{\mathrm G,i}
        -
        (1-m_i)\tilde x_{t,i}
    }{
        \Delta_i(t)
    }.
    \label{eq:appendix_directional_recovery}
\end{equation}
Combining Eqs.~\eqref{eq:graded_target_velocity_form}
and~\eqref{eq:appendix_directional_recovery} gives the exact error
relation
\begin{equation}
    \hat{\tilde u}_{\mathrm G,i}
    -
    \tilde u_{\mathrm G,i}
    =
    \Delta_i(t)
    \left(
        \hat{\tilde u}_i-\tilde u_i
    \right).
    \label{eq:graded_velocity_error_relation}
\end{equation}
Thus, direct graded-target supervision is equivalent to
\begin{equation}
    \mathcal L_{\mathrm G}
    =
    \E\left[
        w(t)
        \sum_i
        \Delta_i(t)^2
        \left(
            \hat{\tilde u}_i-\tilde u_i
        \right)^2
    \right].
    \label{eq:graded_loss_weighting}
\end{equation}
The same recovered-velocity error therefore receives a different penalty
depending on $m_i$ and $t$.

\noindent\textbf{The gains receive gradients.}
Because the target is not detached, the per-sample derivative of the
direct loss is
\begin{equation}
    \left.
    \frac{\partial \ell_{\mathrm G,i}}
         {\partial m_i}
    \right|_{\hat{\tilde u}_{\mathrm G,i}}
    =
    2w(t)
    \left(
        \hat{\tilde u}_{\mathrm G,i}
        -
        \tilde u_{\mathrm G,i}
    \right)
    \tilde\epsilon_i.
    \label{eq:direct_gain_partial_gradient}
\end{equation}
Equivalently,
\begin{equation}
    \frac{\partial \ell_{\mathrm G,i}}
         {\partial \phi_i}
    =
    2w(t)
    \left(
        \hat{\tilde u}_{\mathrm G,i}
        -
        \tilde u_{\mathrm G,i}
    \right)
    \tilde\epsilon_i\,
    m_i(1-m_i).
\end{equation}
This gradient is generally nonzero, but its sign depends on the current
prediction residual. The issue with direct supervision is therefore not
that the gains cannot learn, but that the objective admits a
parameterization-scale shortcut.

\noindent\textbf{A parameterization-scale shortcut.}
Consider any fixed recovered-velocity predictor
$\bar{\tilde u}_i(\vx_t,t)$. For every value of $m_i$, define the
corresponding graded prediction
\begin{equation}
    \hat{\tilde u}_{\mathrm G,i}^{(m_i)}
    =
    (1-m_i)\tilde x_{t,i}
    +
    \Delta_i(t)\bar{\tilde u}_i.
    \label{eq:equivalent_graded_predictor}
\end{equation}
Substituting Eq.~\eqref{eq:equivalent_graded_predictor} into the recovery
map always yields the same predictor:
\begin{equation}
    \frac{
        \hat{\tilde u}_{\mathrm G,i}^{(m_i)}
        -
        (1-m_i)\tilde x_{t,i}
    }{
        \Delta_i(t)
    }
    =
    \bar{\tilde u}_i.
\end{equation}
Hence, changing $m_i$ along this family does not change the recovered
transport field. Nevertheless, the direct graded-target loss becomes
\begin{equation}
    \ell_{\mathrm G,i}(m_i)
    =
    w(t)\,
    \Delta_i(t)^2
    \left(
        \bar{\tilde u}_i-\tilde u_i
    \right)^2.
\end{equation}
Its derivative along this equivalent-predictor family is
\begin{equation}
    \frac{\mathrm d\ell_{\mathrm G,i}}
         {\mathrm d m_i}
    =
    2w(t)\,t\,\Delta_i(t)
    \left(
        \bar{\tilde u}_i-\tilde u_i
    \right)^2
    \geq 0.
    \label{eq:graded_scale_shortcut}
\end{equation}
Therefore, whenever $t>0$ and the recovered velocity has nonzero error,
decreasing $m_i$ reduces the graded-target loss while leaving the
recovered velocity unchanged. The direct objective can thus improve by
making the target more $\vx_0$-like, rather than by learning a more
accurate transport field.

This does not imply that every instantaneous minibatch gradient must
decrease $m_i$. Instead, Eq.~\eqref{eq:graded_scale_shortcut} shows that
the joint optimization problem contains a loss-reducing direction that is
unrelated to transport accuracy.

For the binary AsymFlow parameterization,
\begin{equation}
    \Delta_i(t)^2
    =
    \begin{cases}
        1,       & m_i=1,\\
        (1-t)^2, & m_i=0.
    \end{cases}
\end{equation}
Direct $\vu_{\mathrm G}$ supervision therefore suppresses
recovered-velocity errors in the clean-image-prediction subspace by
$(1-t)^2$. Near the clean endpoint, $t\rightarrow1$, these errors become
almost unpenalized, even though they are amplified when the prediction is
converted back to velocity. 

\noindent\textbf{Velocity-space supervision.}
\method instead applies the standard flow-matching objective after exact
velocity recovery:
\begin{equation}
    \mathcal L_{\vu}(\vtheta,\vphi)
    =
    \E\left[
        w(t)
        \sum_i
        \left(
            \hat{\tilde u}_i-\tilde u_i
        \right)^2
    \right].
    \label{eq:appendix_velocity_loss}
\end{equation}
Using Eq.~\eqref{eq:graded_velocity_error_relation}, this is equivalently
\begin{equation}
    \mathcal L_{\vu}
    =
    \E\left[
        w(t)
        \sum_i
        \frac{
            \left(
                \hat{\tilde u}_{\mathrm G,i}
                -
                \tilde u_{\mathrm G,i}
            \right)^2
        }{
            \Delta_i(t)^2
        }
    \right].
\end{equation}
The inverse factor from velocity recovery cancels the artificial
$\Delta_i(t)^2$ scaling of the graded target. In particular, along the
equivalent-predictor family in
Eq.~\eqref{eq:equivalent_graded_predictor},
\begin{equation}
    \frac{\mathrm d}
         {\mathrm d m_i}
    \left[
        w(t)
        \left(
            \bar{\tilde u}_i-\tilde u_i
        \right)^2
    \right]
    =
    0.
\end{equation}
Velocity-space supervision therefore evaluates all parameterizations using
the same transport objective. The gains can still adapt because they alter
the conditioning and finite-capacity approximation problem faced by the
network, but they cannot reduce the objective merely by shrinking the
scale of the regression target.

\subsection{Surprising behaviour LPIPS on the Learned \texorpdfstring{$\mM$-Curve}{M-Curve}}
\label{app:lpips_mcurve}

Figure~\ref{fig:mflow_gain_curve}(a) shows that LPIPS substantially
changes the prediction parameterization learned by \method. Without
LPIPS, the gains decrease approximately monotonically with PCA variance:
high-variance directions receive more $\vu$-like prediction, while
lower-variance directions become increasingly $\vx_0$-like. This behavior
is consistent with the normalized-variance initialization. With LPIPS,
however, the leading gains become non-monotonic. Using zero-based
indexing, PC0 and PC1 retain high gains, PC2--PC3 exhibit a pronounced
dip, and PC4 rises again.

\begin{figure}[t]
    \centering
    \includegraphics[width=\linewidth]{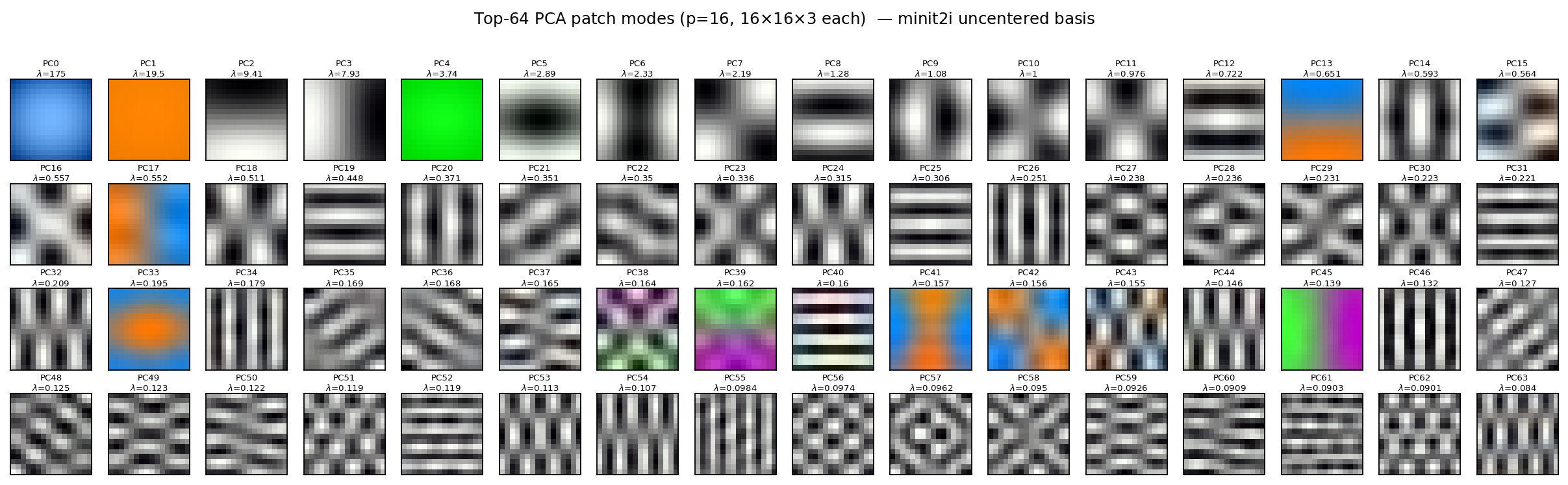}
    \vspace{-0.7em}
    \caption{
        \textbf{Leading modes of the uncentered patch-PCA basis.}
        We reshape the first 64 eigenvectors of the
        \(16\times16\times3\) patch-PCA basis into RGB patches; the value
        \(\lambda\) denotes the variance of each direction. The leading
        ordering interleaves patch-global or very low-frequency modes
        with spatially varying modes. In particular, PC0 is dominated by
        low-frequency patch-level tone, PC1 is an approximately uniform
        chromatic mode, PC2--PC3 encode spatial gradients, and PC4 returns
        to an approximately uniform chromatic mode. Eigenvector signs are
        arbitrary, so the interpretation is based on spatial structure
        rather than the displayed color polarity.
    }
    \label{fig:pca_modes_lpips}
    \vspace{-0.8em}
\end{figure}

The PCA modes in Fig.~\ref{fig:pca_modes_lpips} provide a structural
interpretation of this pattern. PC0 represents a patch-global,
low-spatial-frequency component, while PC1 is nearly constant across the
patch and primarily changes its color. In contrast, PC2 and PC3 contain
vertical and horizontal spatial gradients. PC4 then returns to a nearly
spatially uniform chromatic component. The leading sequence is therefore
approximately
\[
    \text{patch-global},\quad
    \text{patch-global},\quad
    \text{spatial},\quad
    \text{spatial},\quad
    \text{patch-global},
\]
which closely matches the learned gain pattern
\[
    \text{high},\quad
    \text{high},\quad
    \text{low},\quad
    \text{low},\quad
    \text{high}.
\]
In particular, PC4 has substantially lower variance than PC2 and PC3
(\(\lambda_4=3.74\), compared with
\(\lambda_2=9.41\) and \(\lambda_3=7.93\)), yet LPIPS assigns it a
larger gain. The spike at PC4 is therefore not explained by variance
ordering alone; it coincides with the return of a patch-global chromatic
mode after two spatial-gradient modes.

This effect is also more specific than a preference for color. Several
later PCA directions contain strong chromatic variation but are spatially
non-uniform, and they do not receive the same gain increase. The observed
pattern instead suggests that LPIPS distinguishes between patch-global
tone or color changes and spatially varying patterns.

One possible explanation follows from the conditional-mean behavior of
squared-error regression. In a finite-capacity model, a more
$\vx_0$-like parameterization can shrink uncertain components toward
their conditional means. For a spatially varying component, such
shrinkage may primarily smooth a local gradient or texture. For a
patch-global component, however, it can coherently alter the brightness,
contrast, or color of the entire patch. Because LPIPS is applied to the
recovered clean image, it may make these perceptually coherent shifts
more costly. The learned parameterization responds by retaining a larger
velocity-like component for the patch-global modes, thereby reducing
their tendency toward mean reversion.

The controlled comparison establishes that adding LPIPS produces the
non-monotonic $\mM$-curve and that the unusually high gains align with
patch-global PCA modes. The mean-reversion explanation is a plausible
interpretation of this alignment rather than a causal mechanism directly
established by the ablation.

\subsection{A Capacity-Constrained View of \texorpdfstring{$\vx_0$}{x0}- and \texorpdfstring{$\vu$}{u}-Prediction}
\label{app:velocity_budget}

The choice between clean-data and velocity prediction involves two
competing considerations. Clean-data prediction avoids asking the model
to reproduce full-dimensional Gaussian noise, which can be advantageous
when a high-dimensional representation has much lower intrinsic
dimension. Velocity prediction, on the other hand, provides better
numerical conditioning near the clean endpoint. Our RAEv2 experiment
illustrates this trade-off particularly clearly.

\noindent\textbf{RAEv2 predominantly favors clean-data prediction.}
In our clean-latent RAEv2 setup, the pure $\vx_0$-prediction baseline
obtains an FID of $1.05$, while AsymFlow with $r=8$ and MFlow obtain
$1.08$ and $1.09$, respectively, after 80 epochs
(Table~\ref{tab:imagenet_rae}). Consistently, the learned MFlow gains in
Fig.~\ref{fig:mflow_gain_curve}(c) move almost entirely toward zero,
recovering an approximately pure $\vx_0$ parameterization. These results
indicate that, for this representation, model, and training budget, the
network prefers to avoid predicting noise in most latent directions.

At the same time, the AsymFlow result is only $0.03$ FID behind the
pure-$\vx_0$ baseline, despite using full velocity prediction in its
leading eight PCA directions. This result does not show that velocity
prediction is superior in this setting; pure $\vx_0$-prediction remains
best in Table~\ref{tab:imagenet_rae}. It instead shows that a small
velocity-prediction subspace can be accommodated at little measured cost,
even in a representation that strongly favors $\vx_0$-prediction overall.

\noindent\textbf{Numerical conditioning versus target complexity.}
The benefit of retaining a velocity-like component can be understood from
the recovery map. Under our convention,
\[
    \vx_t=t\vx_0+(1-t)\bm\epsilon,
    \qquad t=1\text{ is the clean endpoint}.
\]
For pure $\vx_0$-prediction, the velocity is recovered as
\begin{equation}
    \hat{\tilde u}_i
    =
    \frac{
        \hat{\tilde x}_{0,i}-\tilde x_{t,i}
    }{
        1-t
    }.
\end{equation}
Consequently, an error
$\delta_i^{x}=\hat{\tilde x}_{0,i}-\tilde x_{0,i}$ in the clean-data
prediction produces a velocity error
\begin{equation}
    \hat{\tilde u}_i-\tilde u_i
    =
    \frac{\delta_i^{x}}{1-t}.
    \label{eq:xpred_velocity_amplification}
\end{equation}
The factor $1/(1-t)$ becomes ill-conditioned as $t\rightarrow1$.

For MFlow, the corresponding relation is
\begin{equation}
    \hat{\tilde u}_i-\tilde u_i
    =
    \frac{
        \hat{\tilde u}_{\mathrm G,i}
        -
        \tilde u_{\mathrm G,i}
    }{
        \Delta_i(t)
    },
    \qquad
    \Delta_i(t)=(1-t)+tm_i.
    \label{eq:mflow_velocity_amplification}
\end{equation}
When $m_i=0$, this reduces to the same $1/(1-t)$ amplification as
$\vx_0$-prediction. For any $m_i>0$, however,
\[
    \lim_{t\rightarrow1}\Delta_i(t)=m_i,
\]
so the clean-endpoint singularity is removed in that direction. Thus,
conditional on comparable target-prediction errors, a nonzero velocity
component gives better-conditioned velocity recovery near clean samples.

This conditioning advantage comes with an opposing cost. Increasing
$m_i$ changes the target from clean-data prediction toward
\[
    \tilde u_{\mathrm G,i}
    =
    \tilde x_{0,i}-m_i\tilde\epsilon_i,
\]
requiring the finite-capacity network to retain a larger Gaussian-noise
component. Full velocity prediction is therefore well conditioned during
recovery but can be difficult to approximate in high-dimensional spaces,
whereas pure $\vx_0$-prediction removes the noise-prediction burden but
retains the clean-endpoint amplification. The optimal parameterization
must balance these two effects.

\paragraph{A soft velocity budget.}
This trade-off suggests viewing MFlow as learning both \emph{how much}
velocity prediction the model can support and \emph{where} it should be
allocated. A useful summary of the former is the soft velocity rank
\begin{equation}
    r_{\mathrm{eff}}
    \coloneqq
    \operatorname{tr}(\mM)
    =
    \sum_i m_i.
\end{equation}
Pure $\vx_0$-prediction has $r_{\mathrm{eff}}=0$, full
$\vu$-prediction has $r_{\mathrm{eff}}=D$, and rank-$r$ AsymFlow has
$r_{\mathrm{eff}}=r$. For MFlow, this quantity is continuous and learned.

Our results suggest a capacity-constrained principle:
\emph{use velocity-like prediction in as many directions as the model can
accurately support, and retain clean-data-like prediction in the remaining
directions}. The RAEv2 curve indicates a small velocity budget for this
high-dimensional latent representation, whereas the text-to-image curves
show that increasing model capacity shifts more directions toward
velocity prediction. We therefore interpret model capacity as influencing
the available velocity budget, while the data representation and training
objective determine how that budget should be distributed.

\noindent\textbf{Why leading PCA directions are a natural default.}
Suppose that only $N$ directions can be assigned substantial velocity
gains. In the absence of perceptual or other auxiliary losses, the leading
PCA directions provide a natural initial allocation. In the PCA basis,
\begin{equation}
    \operatorname{Var}(\tilde x_{0,i})=\lambda_i,
    \qquad
    \operatorname{Var}(\tilde\epsilon_i)=1,
\end{equation}
where $\lambda_i$ decreases with $i$, while the isotropic noise variance
is identical in every direction. The velocity target therefore becomes
increasingly noise-dominated as $\lambda_i$ decreases. Allocating a
limited velocity budget to high-variance directions concentrates the
noise-prediction burden in the dominant data subspace, while keeping the
many low-variance, predominantly off-manifold directions closer to
$\vx_0$-prediction. It also improves near-clean conditioning in the
directions carrying the largest fraction of the data variation.

This PCA ordering should nevertheless be viewed as an initialization
prior rather than a universal rule. Auxiliary objectives can favor
directions according to their visual function rather than variance, as
shown by the non-monotonic LPIPS $\mM$-curve in
Fig.~\ref{fig:mflow_gain_curve}(a). We also observe that a non-leading PCA
block can outperform the first eight directions in one ImageNet pilot.
MFlow therefore initializes gains using normalized variance but lets the
model modify both the effective velocity budget and its direction-wise
allocation through end-to-end training.

\subsection{Radially-Averaged Power Spectrum}
\label{sec:raps}
To quantify how faithfully a model reproduces the spatial-frequency content of natural images, we compare the azimuthally-averaged 2D power spectrum of generated and real images. Each image is converted to luma ($Y=0.299R+0.587G+0.114B$), mean-subtracted to remove the DC component, and multiplied by a separable 2D Hann window to suppress boundary leakage. We compute the squared magnitude of its 2D DFT, $P(\mathbf{f})=|\mathcal{F}{g}(\mathbf{f})|^2$, center it (fftshift), and azimuthally average it, \ie, average over all orientations at each fixed radius $r=\lVert\mathbf{f}\rVert$, to obtain a 1D profile $P(f)$ as a function of radial frequency $f$ (cycles/pixel). Averaging $P(f)$ over $N=4000$ images gives the RAPS of a distribution.

\subsection{Refiners}
\label{sec:refiner_more_details}

For our \emph{ConvRefiner} we use a $t$-conditioned PatchGAN discriminator (\cite{isola2017image}) (NLayerDiscriminator, 3 layers, ndf=64) with a hinge adversarial loss, applied to the refined $\hat{\vx}_0$. We have two types of batches -- all-$t$ batch and near-clean batch ($t\sim\mathcal{U}(0.9,0.98)$). For the all-$t$ batch the flow-matching and LPIPS($t\ge0.3$) losses are normally applied. For the near-clean batch, additional GAN loss is applied with a $0.02$ weight, if not otherwise specified, and the flow-matching and LPIPS loss are downweighted by a factor of $0.25$.  For Attention Refiner we used a 2 blocks of swin-style (\cite{liu2021swin}) windowed attention residual refiner in the clean image space, as a drop in replacement for the \emph{ConvRefiner}. Note that DiP patch-detailer also takes as input $\vx_t$ apart from the model output.

\begin{table*}[t!]
    \centering

    \begin{minipage}[t]{0.48\linewidth}
        \vspace{0pt}
        \centering
        \caption{
            \textbf{Learnable gains on UCF-101 (B-level).}
            Results use the default sampling hyperparameters, without an
            evaluation sweep.
        }
        \label{tab:ablation_video_fixed_learnable}
        \vspace{-0.7em}

        \small
        \setlength{\tabcolsep}{5pt}
        \begin{tabular}{@{}lc@{}}
            \toprule
            \textbf{Method} & \textbf{FVD$\downarrow$} \\
            \midrule
            Video \method, fixed gains
            & 169.4 \\
            \rowcolor{gray!20}
            Video \method, learnable gains
            & \textbf{155.5} \\
            \bottomrule
        \end{tabular}
    \end{minipage}
    \hfill
    \begin{minipage}[t]{0.48\linewidth}
        \vspace{0pt}
        \centering
        \caption{
            \textbf{RAEv2 results.}
            H-level models trained on ImageNet for 80 epochs.
        }
        \label{tab:imagenet_rae}
        \vspace{-0.7em}

        \small
        \setlength{\tabcolsep}{5pt}
        \begin{tabular}{@{}lc@{}}
            \toprule
            \textbf{Method} & \textbf{FID$\downarrow$} \\
            \midrule
            RAEv2~\cite{singh2026improved}
            & \textbf{1.05} \\
            RAEv2 + Asym8
            & 1.08 \\
            \rowcolor{gray!20}
            RAEv2 + \method
            & 1.09 \\
            \bottomrule
        \end{tabular}
    \end{minipage}

    \vspace{-1em}
\end{table*}

\begin{figure}[h]
\begin{center}
\vspace{-2em}
\includegraphics[width=0.9\linewidth]{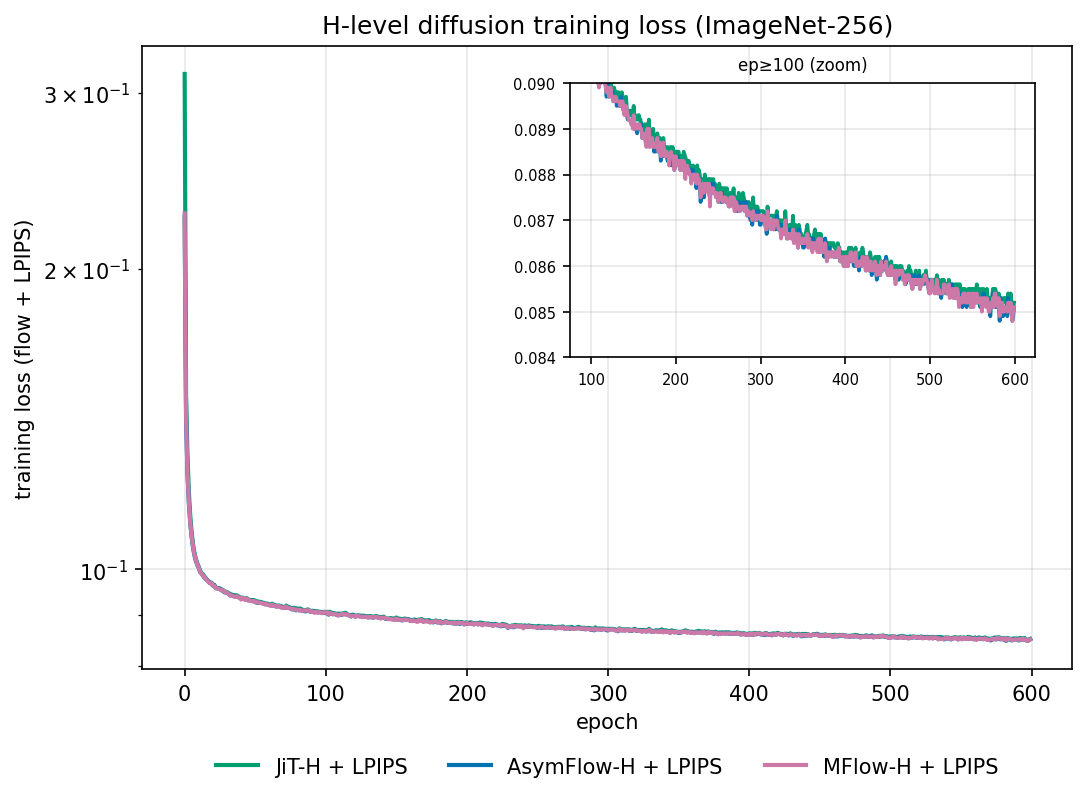}
\end{center}
\vspace{-2em}
\caption{\textbf{ImageNet Loss Curves at H-level.} We usually observe that the $\vu$-loss of \method training is very slightly better.}
\label{fig:imagenet_qual_uncurated2}
\vspace{-1em}
\end{figure}

\begin{figure}[h]
\begin{center}
\includegraphics[width=0.9\linewidth]{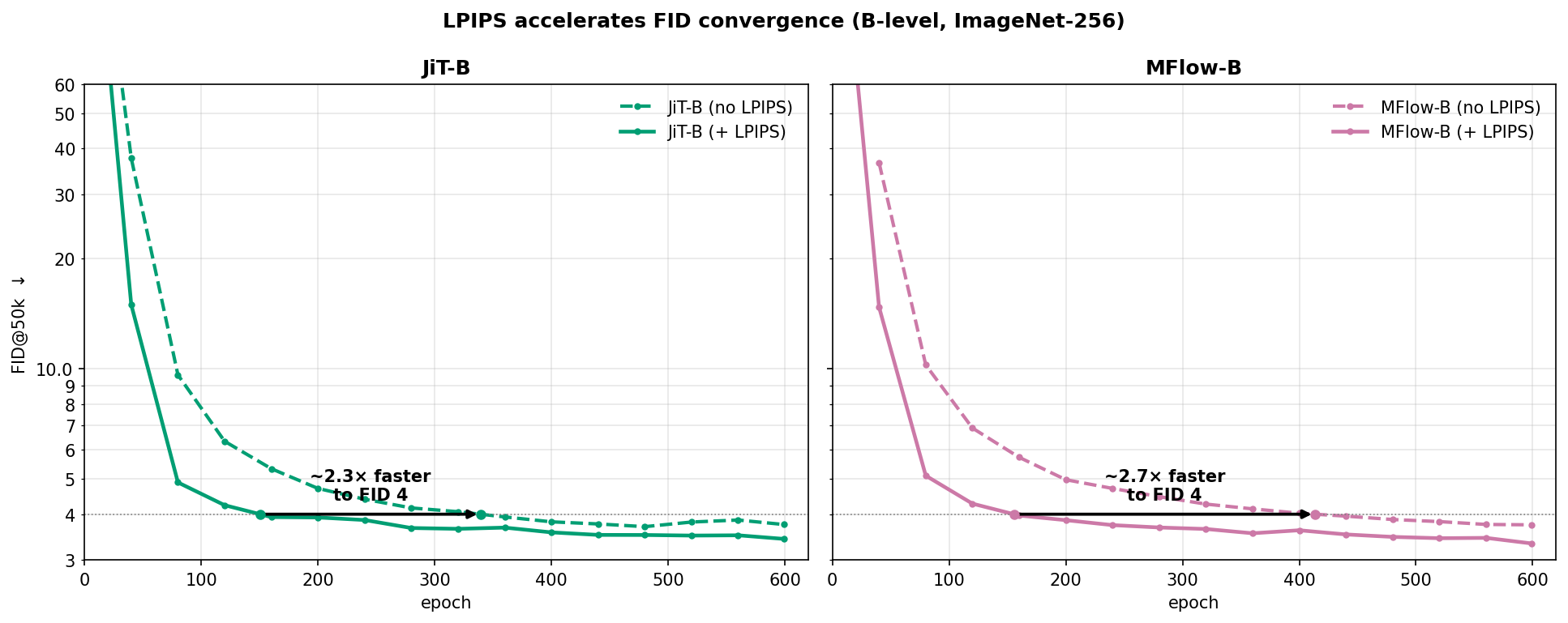}
\end{center}
\vspace{-1em}
\caption{\textbf{Faster convergence on ImageNet with LPIPS at B-level.} This corresponds to the results noted in Table~\ref{tab:imagnet_b}.}
\vspace{-1em}
\label{fig:imagenet_qual_uncurated2}
\end{figure}

\subsection{Limitations.}
Although \method is a general formulation that transfers across
pixel-space image and video generation, high-dimensional representation
latents such as RAEv2~\cite{singh2026improved}, and text-to-image models
of different capacities, its improvements over strong fixed
parameterizations are sometimes modest. One possible reason is the joint
optimization of the gains and network weights. While the underlying
velocity target remains fixed, updating $\mM$ continuously changes the
network's direct output parameterization
$\vu_{\mathrm G}=\vx_0-\mM\bm\epsilon$, creating an effectively moving
regression target and allowing the gains and network to co-adapt. Indeed,
under the same velocity-space objective, \method attains a slightly lower
training loss than JiT and AsymFlow on the ImageNet H-level experiment
(Fig.~\ref{fig:imagenet_qual_uncurated2}); we observe similar trends for
B-level models with and without LPIPS. However, the lower diffusion loss
does not consistently translate into better FID. This mismatch suggests
that improved regression conditioning need not directly improve
distribution-level sample quality, and its precise cause remains unclear.
Staged or decoupled optimization of the gains is an important direction
for future work.

\emph{ConvRefiner} reveals a complementary evaluation limitation. It
substantially closes the high-frequency spectral gap and brings the
generated spectrum closer to that of real images, yet this improvement
produces little or no change in FID. This suggests that FID is relatively
insensitive to the local frequency statistics modified by the refiner.
Consequently, spectral fidelity and FID should be viewed as complementary
rather than interchangeable measures; texture-sensitive metrics and
human evaluation are needed to determine how strongly the recovered
high-frequency content translates to perceived image quality.

\newpage

\begin{figure}[h]
\begin{center}
\includegraphics[width=\linewidth]{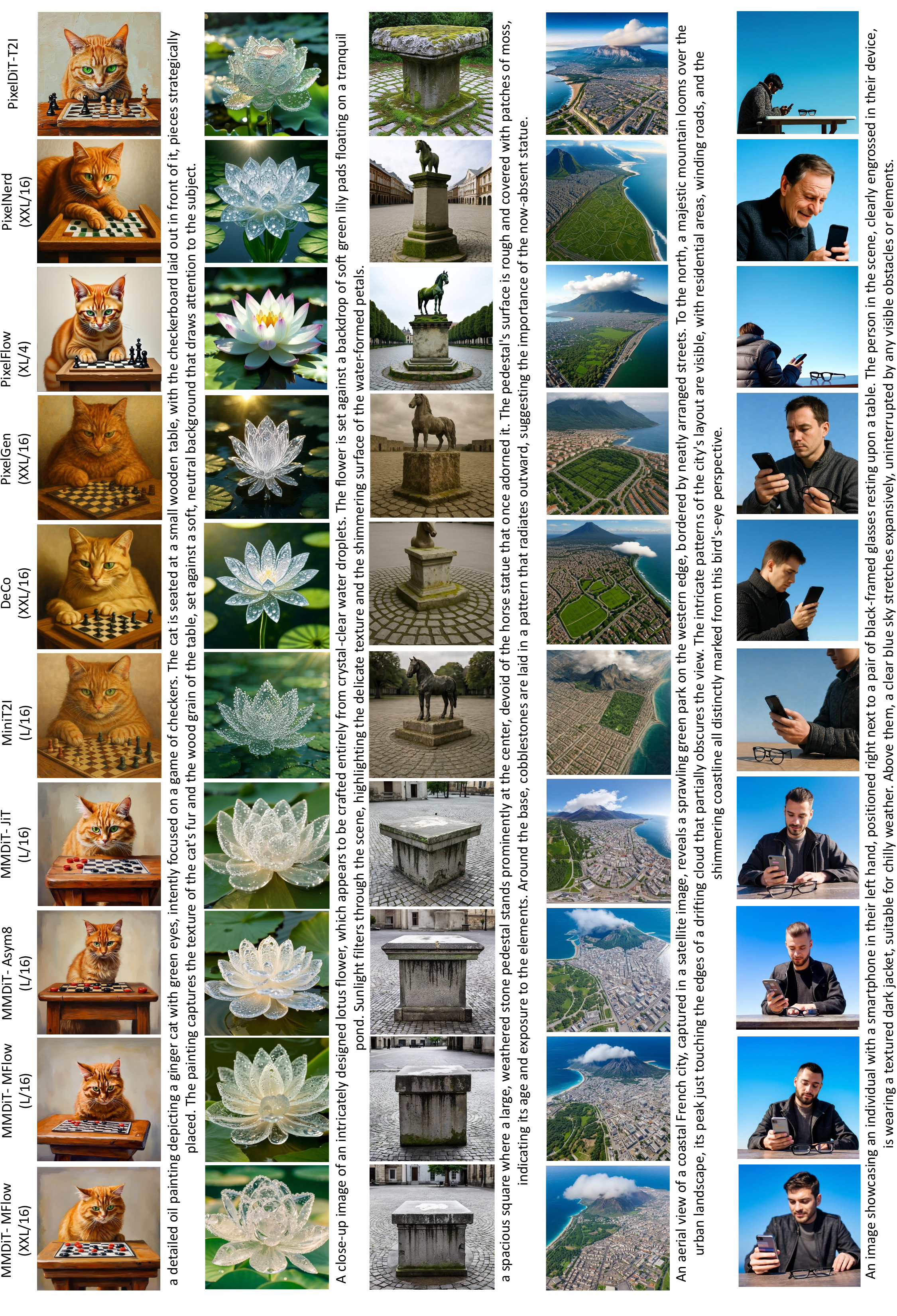}
\end{center}
\caption{\textbf{Qualitative Text-to-Image Results on DPG-Bench.} Images generated by our method and baselines on DPG-Bench prompts.}
\label{fig:imagenet_qual}
\end{figure}

\begin{figure}[h]
\begin{center}
\includegraphics[width=\linewidth]{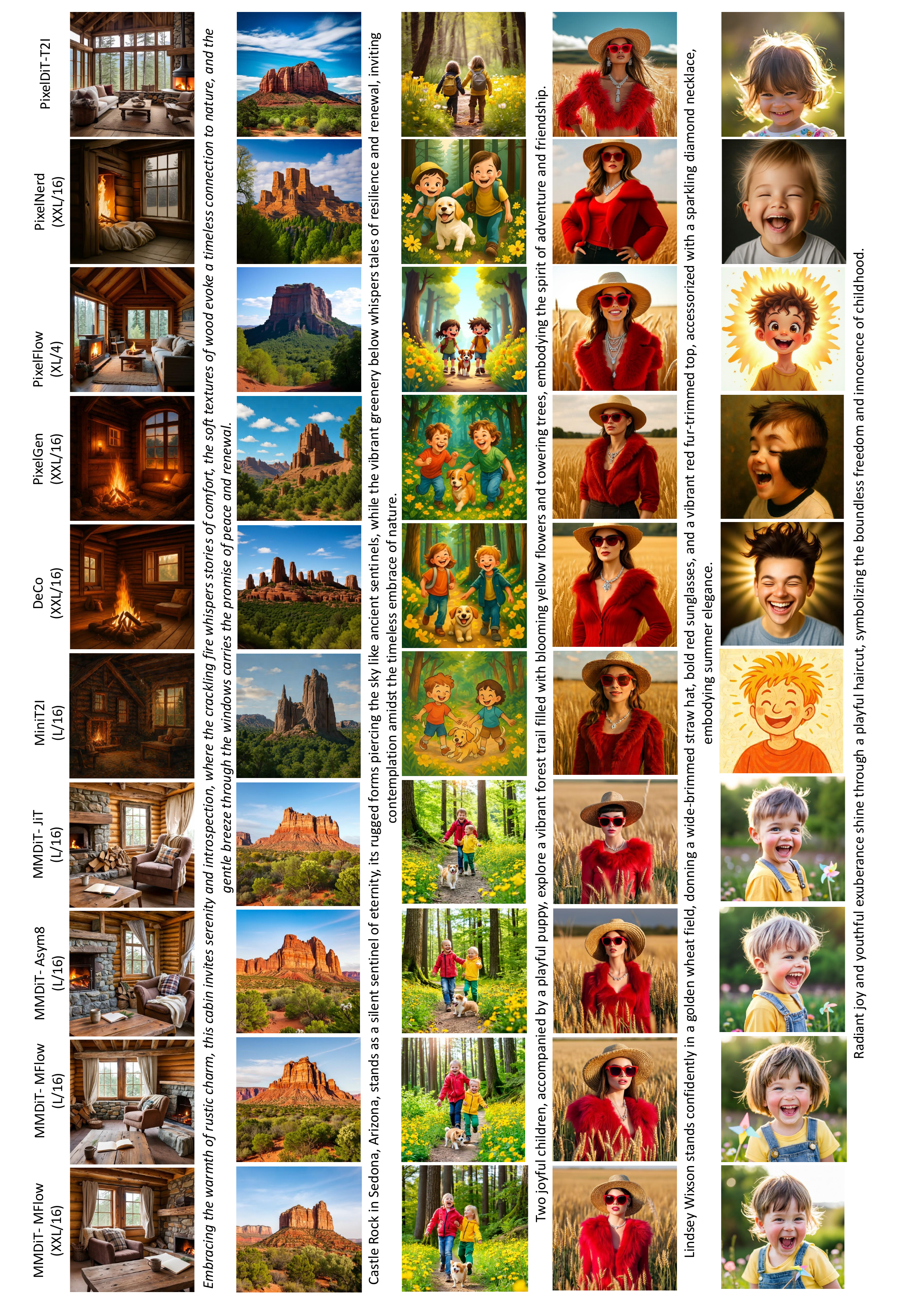}
\end{center}
\caption{\textbf{Qualitative Text-to-Image Results on PRISM-Bench.} Images generated by our method and baselines on PRISM-Bench prompts.}
\label{fig:imagenet_qual}
\end{figure}

\begin{figure}[h]
\begin{center}
\includegraphics[width=0.9\linewidth]{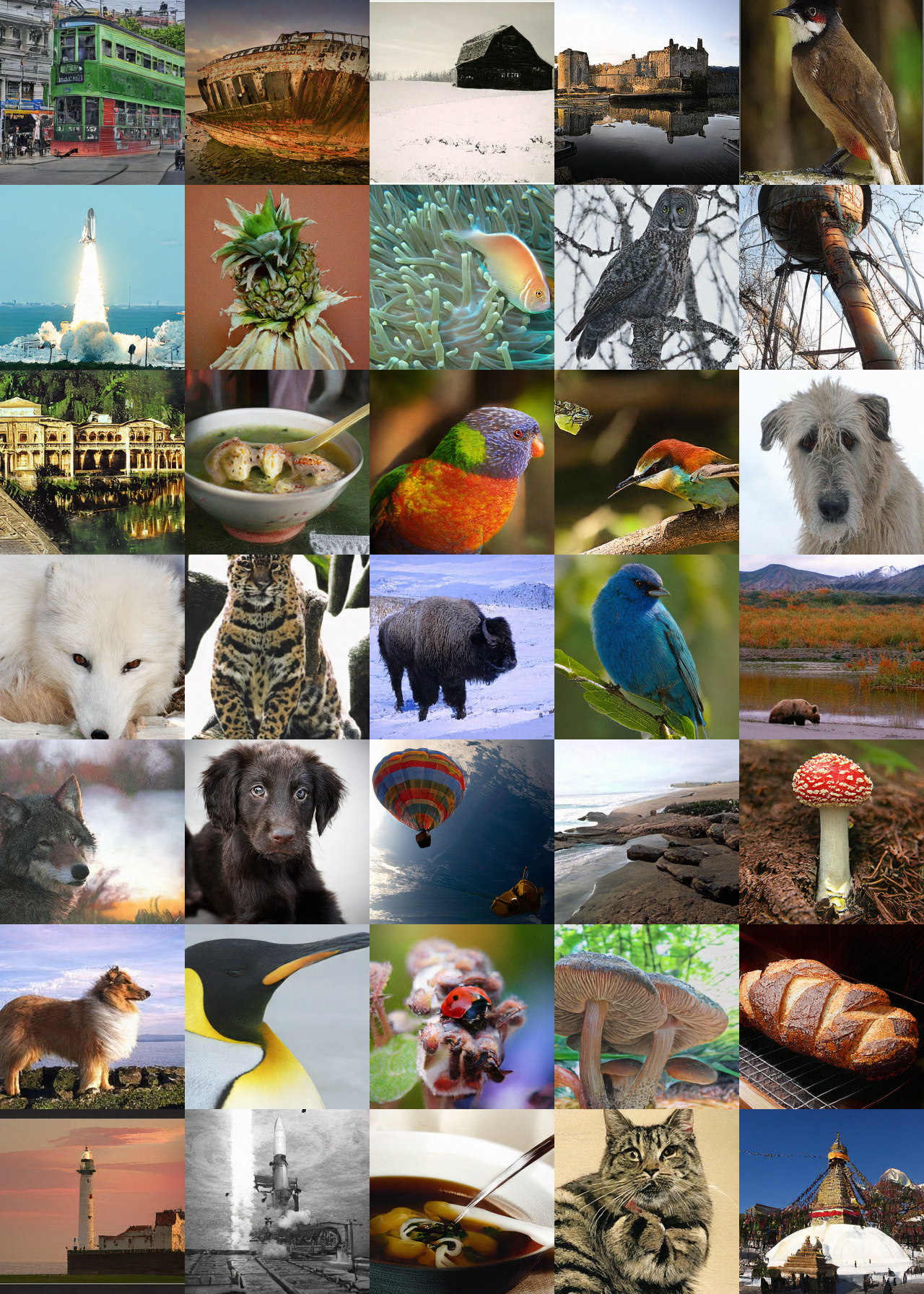}
\end{center}
\caption{\textbf{Qualitative ImageNet Results.} Images generated by \method+\emph{ConvRefiner} H-level model.}
\label{fig:imagenet_qual}
\end{figure}

\begin{figure}[h]
\begin{center}
\includegraphics[width=0.9\linewidth]{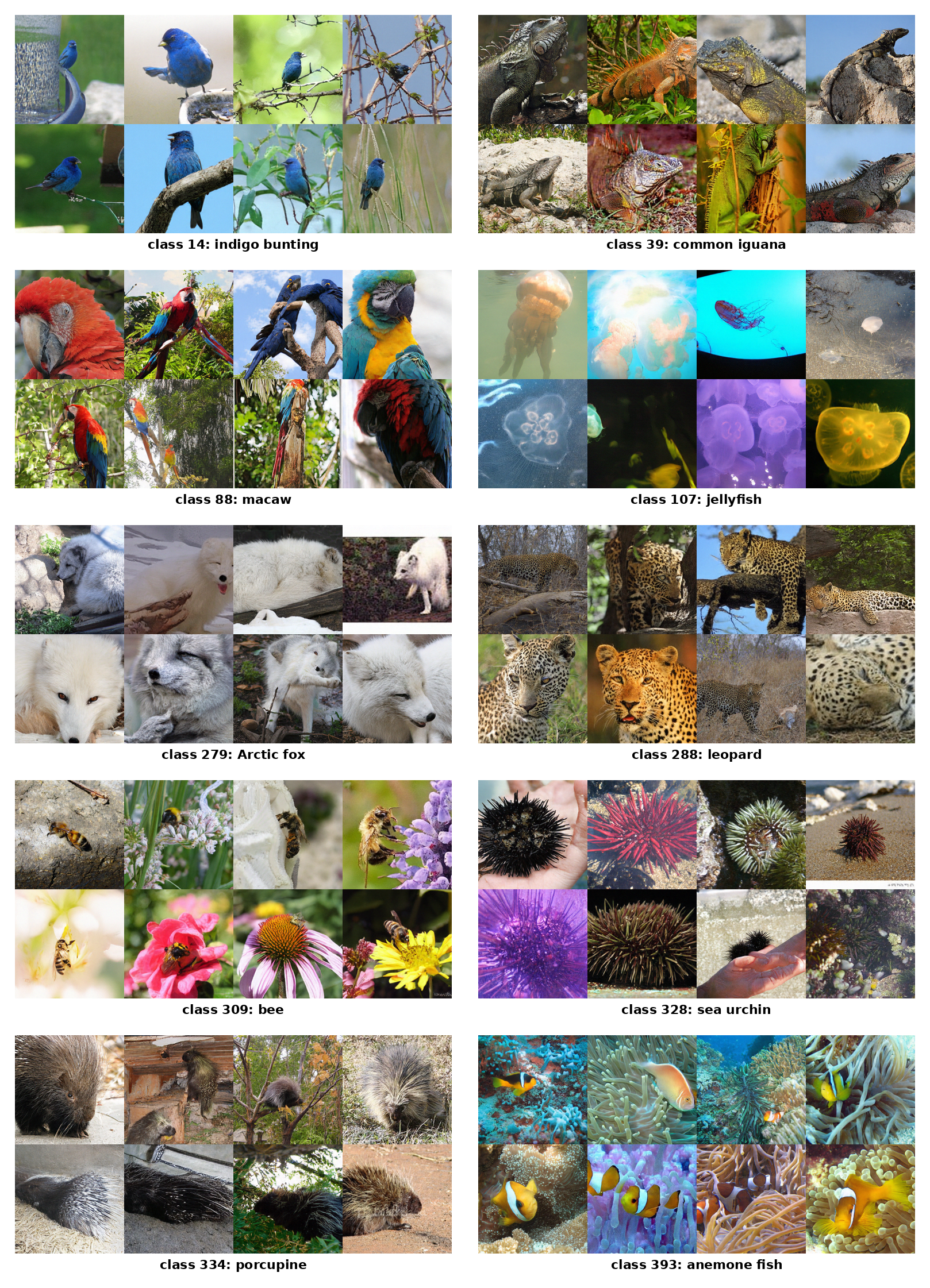}
\end{center}
\caption{\textbf{More Class-wise Uncurated Qualitative ImageNet Results.} Images generated by \method+\emph{ConvRefiner} H-level model.}
\label{fig:imagenet_qual_uncurated1}
\end{figure}

\begin{figure}[h]
\begin{center}
\includegraphics[width=0.9\linewidth]{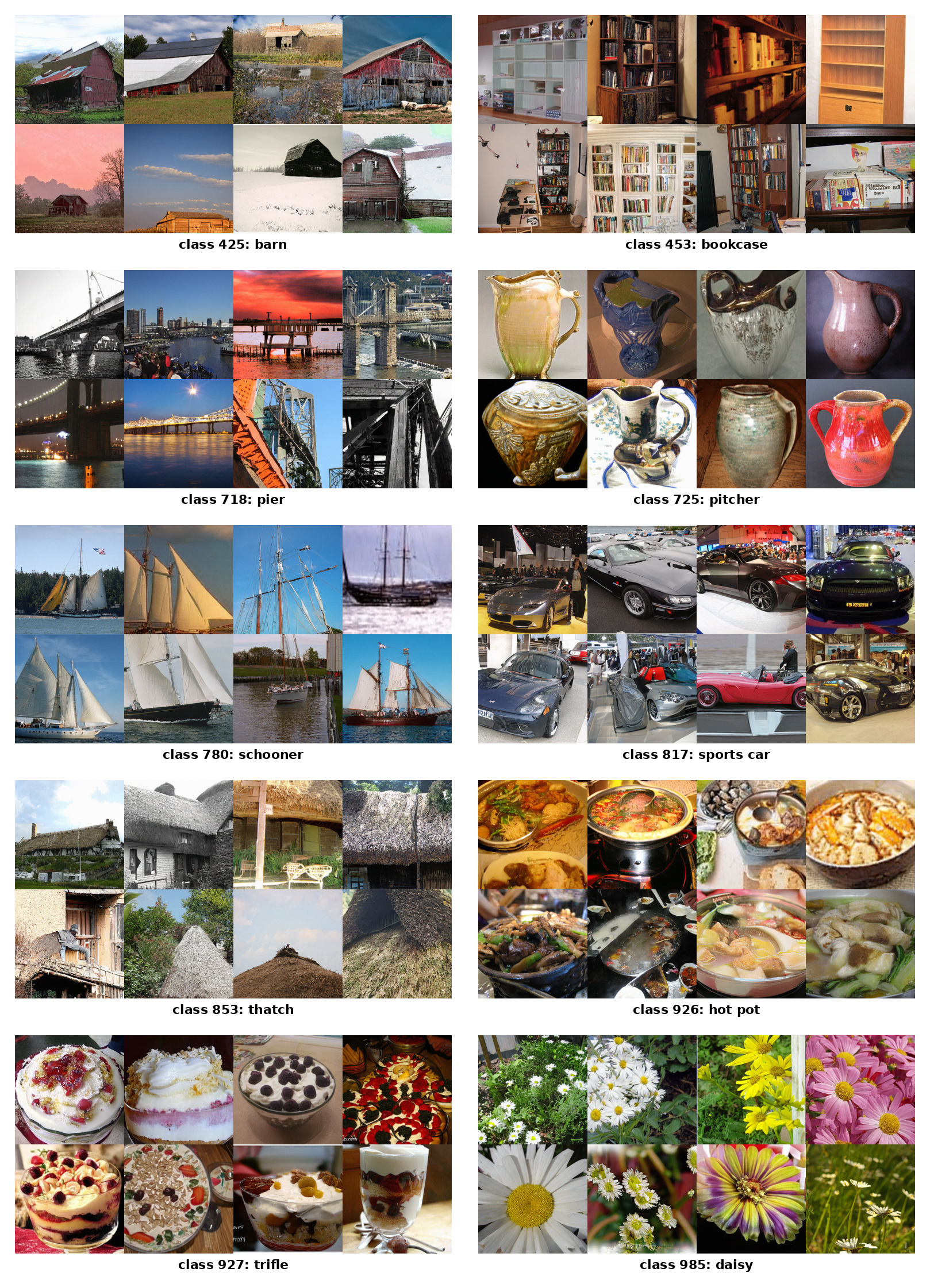}
\end{center}
\caption{\textbf{More Class-wise Uncurated Qualitative ImageNet Results.} Images generated by \method+\emph{ConvRefiner} H-level model.}
\label{fig:imagenet_qual_uncurated2}
\end{figure}

\begin{figure}[h]
\begin{center}
\includegraphics[width=0.9\linewidth]{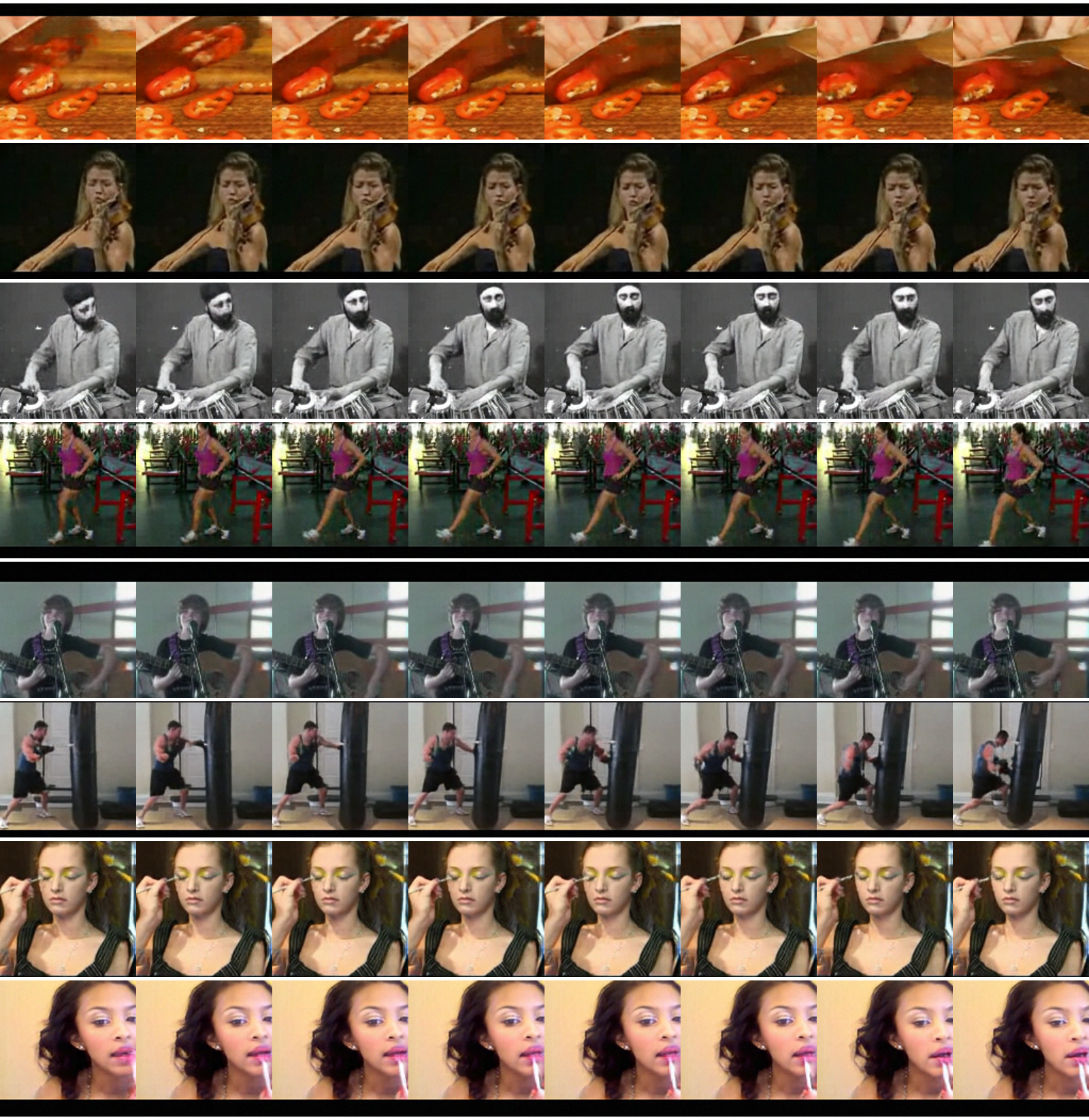}
\end{center}
\caption{\textbf{Qualitative UCF-101 Results.} Frames generated by \method+post-hoc FW-LPIPS L-level model.}
\label{fig:imagenet_qual_uncurated2}
\end{figure}

\begin{figure}[h]
\begin{center}
\includegraphics[width=0.8\linewidth]{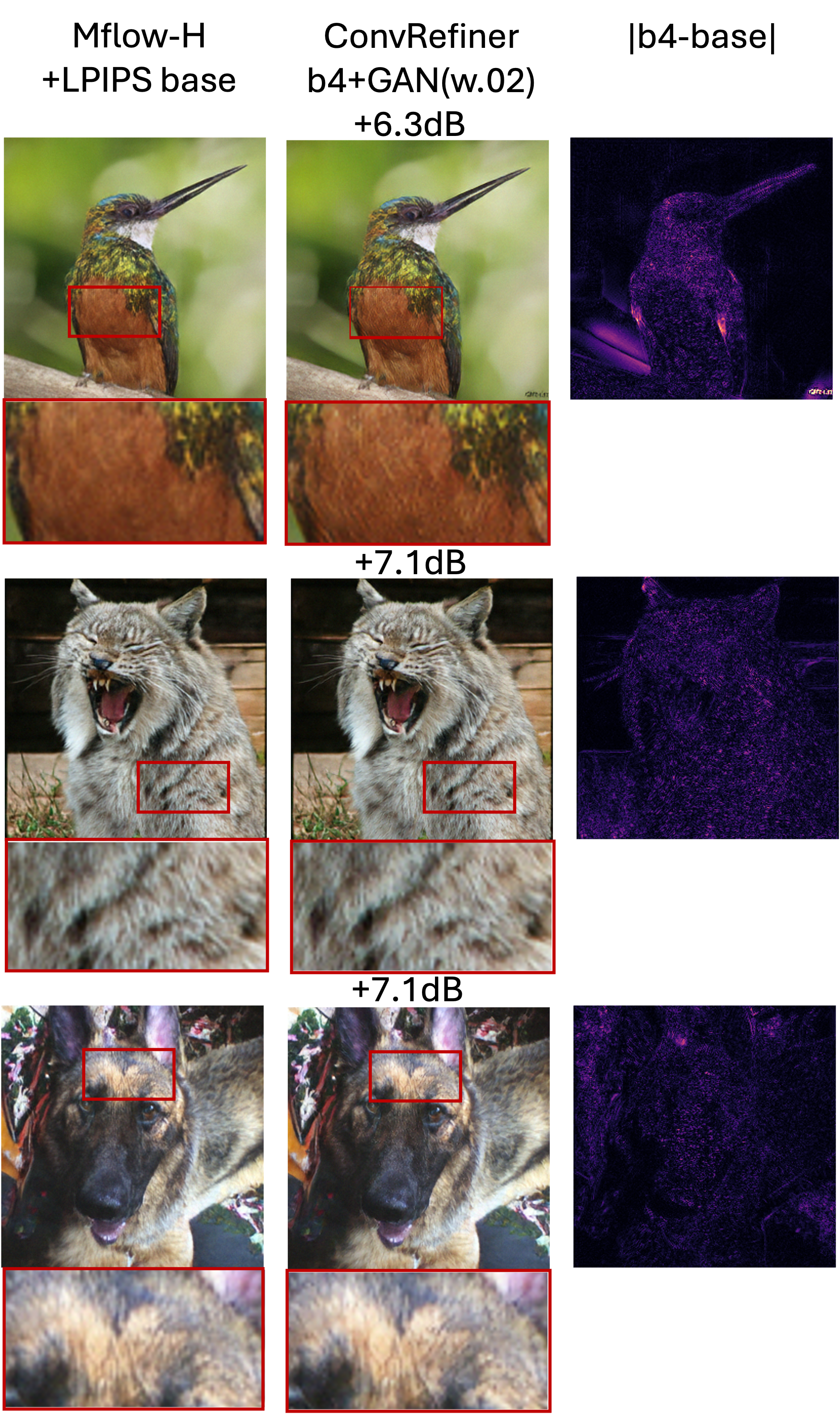}
\end{center}
\caption{\textbf{Visualizing effect of \emph{ConvRefiner}.} Here we visualize the effect of adding our post-training to \method-H+LPIPS on ImageNet generations. We also note the average power gain in the high-frequency(0.5–0.7) region. We see overall improvement in texture details.}
\label{fig:convrefiner_qual}
\end{figure}

\begin{figure}[h]
\begin{center}
\includegraphics[width=0.9\linewidth]{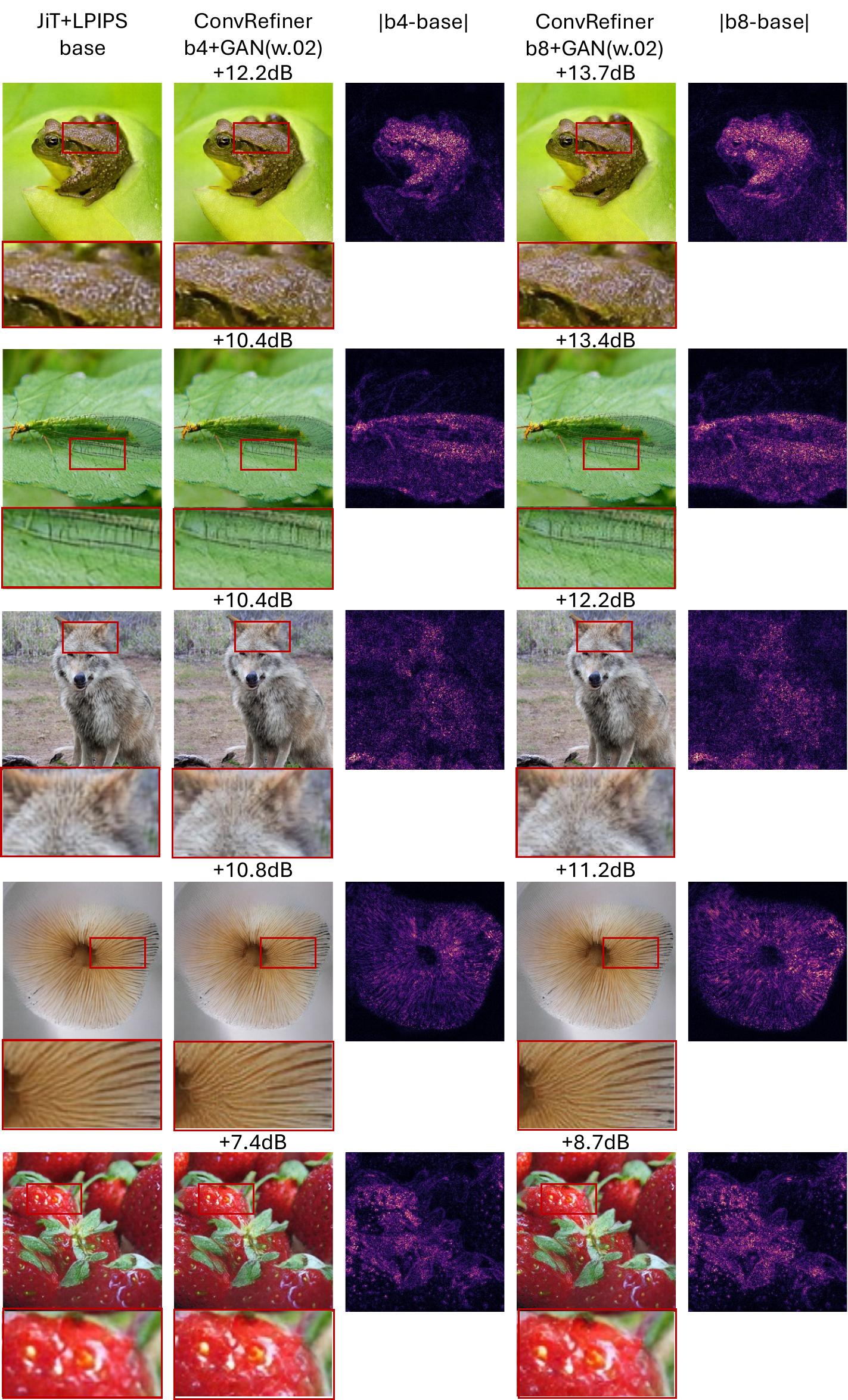}
\end{center}
\caption{\textbf{Qualitative Ablation for \emph{ConvRefiner}.} Here we visualize the effect of adding our post-training to JiT-B on ImageNet generations. We also note the average power gain in the high-frequency(0.5–0.7) region. We also show the visual improvements -- we see that b4 adds additional texture details, while b8 sometimes overshoots in texture.  }
\label{fig:convrefiner_ablation_qual}
\end{figure}

\end{document}